\documentclass[10pt, a4paper]{article}

\usepackage{lrec-coling2024} 
\usepackage{inconsolata}
\usepackage{float}
\usepackage{graphicx}
\usepackage{amsfonts,amssymb}
\usepackage{booktabs}
\usepackage{multirow}
\usepackage{makecell} 
\usepackage{bm}
\usepackage{amsmath}
\usepackage{colortbl}

\title{Zero-Shot Cross-Lingual Document-Level Event Causality Identification with Heterogeneous Graph Contrastive Transfer Learning}

\name{Zhitao He${}^{1,2}$, Pengfei Cao${}^{1,2}$, Zhuoran Jin${}^{1,2}$, Yubo Chen${}^{1,2}$ \\
        {\bf \large Kang Liu${}^{1,2,3}$, Zhiqiang Zhang${}^4$, Mengshu Sun${}^4$, Jun Zhao$^{1,2 \ast}$ \thanks{*Corresponding author}}} 

\address{${}^1$ The Laboratory of Cognition and Decision Intelligence for Complex Systems, 
         \\ Institute of Automation, Chinese Academy of Sciences, Beijing, China 
         \\${}^2$  School of Artificial Intelligence, University of Chinese Academy of Sciences, Beijing, China 
         \\${}^3$  Shanghai Artificial Intelligence Laboratory, China
         \\${}^4$  Ant Group, Hangzhou, China
         \\ \texttt{\{zhitao.he, pengfei.cao, yubo.chen, kliu, jzhao\}@nlpr.ia.ac.cn}\\ }

\abstract{
Event Causality Identification (ECI) refers to the detection of causal relations between events in texts. However, most existing studies focus on sentence-level ECI with high-resource languages, leaving more challenging document-level ECI (DECI) with low-resource languages under-explored. In this paper, we propose a Heterogeneous \textbf{G}raph \textbf{I}nteraction Model with \textbf{M}ulti-granularity \textbf{C}ontrastive Transfer Learning (GIMC) for zero-shot cross-lingual document-level ECI. Specifically, we introduce a heterogeneous graph interaction network to model the long-distance dependencies between events that are scattered over a document. Then, to improve cross-lingual transferability of causal knowledge learned from the source language, we propose a multi-granularity contrastive transfer learning module to align the causal representations across languages. Extensive experiments show our framework outperforms the previous state-of-the-art model by 9.4\% and 8.2\% of average F1 score on monolingual and multilingual scenarios respectively. Notably, in the multilingual scenario, our zero-shot framework even exceeds GPT-3.5 with few-shot learning by 24.3\% in overall performance. 
 \\ \newline \Keywords{zero-shot cross-lingual, document-level, event causality identification} }

\begin{document}

\maketitleabstract

\section{Introduction}

Event Causality Identification (ECI) is an important task in natural language processing (NLP), which can facilitate various applications, including explainable question answering \citep{yang2018hotpotqa}, intelligent search \citep{rudnik2019searching} and complex reasoning \citep{dalvi-etal-2021-explaining}. Most previous methods \citep{kadowaki-etal-2019-event, ijcai2020p499, zuo-etal-2021-improving, cao-etal-2021-knowledge} focus on sentence-level with English corpora. Nevertheless, a substantial number of causal relations are expressed by multiple sentences. For instance, there are approximately 68.7\% of causal relationships in English corpora \cite{caselli2017event} are attributed to inter-sentence event pairs. Hence, identifying causality of events at the document-level is necessary, which gains increasing attention recently \citep{tran-phu-nguyen-2021-graph, fan2022towards,MIR-2022-10-329, chen-etal-2022-ergo}. 

\begin{figure}[t]
	 	\centering{ 
	 	\includegraphics[width=0.485\textwidth]{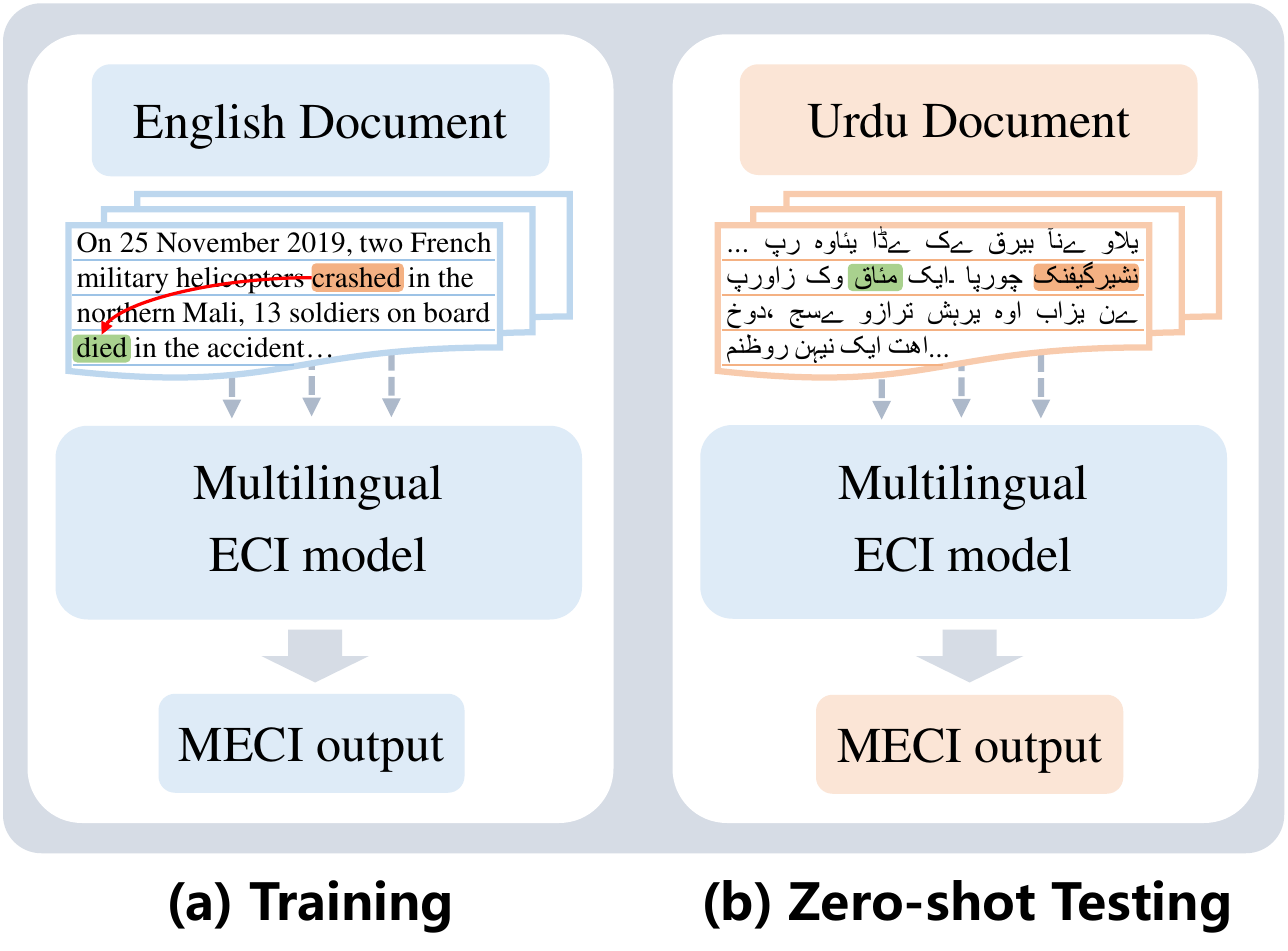}}  
	 	\caption{Zero-shot cross-lingual document-level event causality identification. Data is non-parallel.}
	 	\label{fig2}
\end{figure}

However, training an ECI model typically relies on a large amount of data, especially for document-level, which makes it hard to adapt to low-resource languages. While pre-trained language models have exhibited remarkable capabilities across various tasks \citep{brown2020language, chowdhery2022palm, wang2023cross, tan2023evaluation,MIR-2022-07-221}, they still struggle in multilingual setting, as evidenced in recent studies \citep{chang2023survey, huang2023not, zhang2023m3exam}. Even the powerful ChatGPT exhibits quite limited performance in multilingual relation prediction task, with an average accuracy of only 37\%. Particularly for low-resource languages, the model performs even worse, such as achieving only a 6.3\% accuracy in Urdu \cite{lai2023chatgpt}. Moreover, low-resource languages face a critical shortage of training data, making it challenging to enhance the document-level ECI performance of language models. Therefore, in this paper we focus on zero-shot cross-lingual document-level ECI, aiming to efficiently transfer the causality knowledge from the source languages to any other languages (e.g. low-resource/less-studied languages) under limited language resources. As shown in Figure \ref{fig2}, the model is trained with the annotated data in source language and directly applied to target language. While significant efforts have been made for monolingual setting \cite{tran-phu-nguyen-2021-graph, chen-etal-2022-ergo}, two critical challenges arise when we apply monolingual ECI to zero-shot cross-lingual document-level setting:

(1) \textbf{Language-agnostic causal knowledge alignment}. Each language has its own characteristics. The multilingual ECI models, trained in source language, inevitably tend to learn language-specific knowledge rather than pure language-agnostic knowledge (i.e., causal knowledge). Thus, the trained ECI model may only perform well in source language. For example, different languages have distinct distributions of distance between causal events. According to statistics, the majority distance between causal events in English corpora is 40 words, whereas in Turkish, it is 33 words, and in Danish, it is 23 words. Furthermore, based on our error analysis of vanilla pre-trained langugae model, when we employ model trained on English to test on Danish, more than half (53\%) of the false positive causal events are separated by more than 23 words, which means the ECI model trained in English corpora tend to predict a causal relation between two events that are far away, leading negatively impact on the prediction in Danish.

\begin{figure}[t]
	 	\centering{ 
	 	\includegraphics[width=0.485\textwidth]{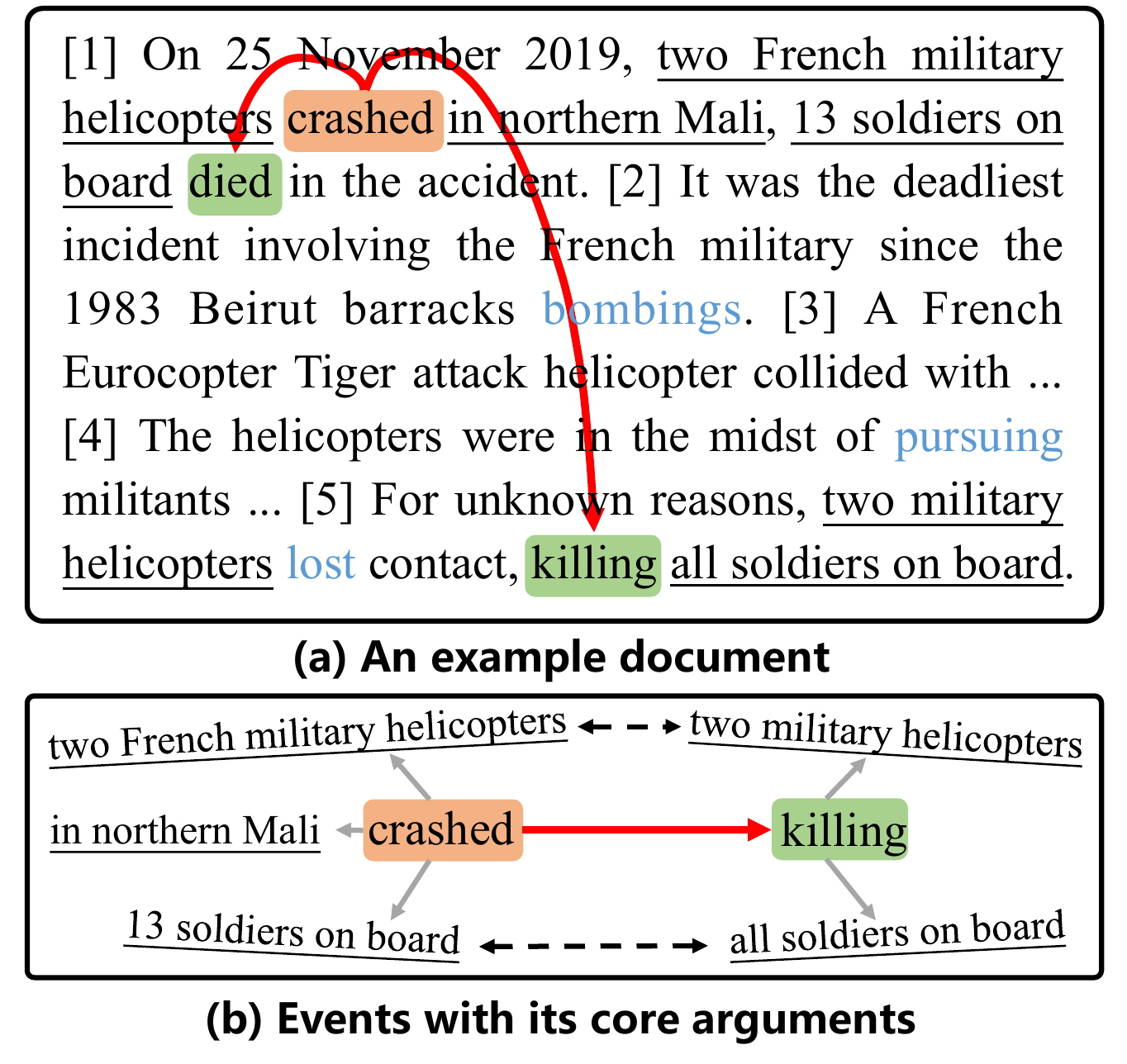}}  
	 	\caption{We show event-relevant sentences. Events are colored and we underline the informative phrases related to colored events. Red lines and gray lines indicate the causal relations and arguments of events, respectively. }
	 	\label{fig1}
\end{figure}

(2) \textbf{Causal events scattering}. 
As shown in Figure \ref{fig1}(a), the intra-sentence causality between “\textit{\rm{crashed}}” and “\textit{\rm{died}}” can be easily predicted. By contrast, event “\textit{\rm{crashed}}” and “\textit{\rm{killing}}” are located in sentence 1 and 5 respectively, the long distance and the interference of irrelevant events such as “\textit{\rm{Beirut barracks bombings}}” and “\textit{\rm{pursuing militants}}” make it difficult to directly model the long-distance dependencies between mentioned events. Fortunately, as shown in Figure \ref{fig1}(b), we can efficiently obtain the informative phrases (i.e., the underlined “\textit{\rm{military helicopter”}} and \textit{“\rm{soldier”}}) related to scattered causal events “\textit{\rm{crashed}}” and “\textit{\rm{killing}}” by processing dependency structure with heuristic rules. We find that these informative phrases are similar and can serve as intermediate bridges for connecting the two events, which indicates the importance of informative phrases in modeling long-distance dependencies between events.

To tackle the aforementioned challenges, we propose a Heterogeneous \textbf{G}raph \textbf{I}nteraction Model with \textbf{M}ulti-granularity \textbf{C}ontrastive Transfer Learning (GIMC). Specifically, we introduce a \textbf{\textit{Multi-granularity Contrastive Transfer Learning Module}} to align causal representations across languages for transferring language-agnostic causal knowledge in statement and aspect levels. Meanwhile, we design a \textbf{\textit{Heterogeneous Graph Interaction network}} with informative phrase nodes, sentence nodes, statement nodes, event pair nodes to model the long-distance dependencies between events. Extensive experiments show that our zero-shot framework outperforms previous models and even exceeds GPT-3.5 with zero/few-shot prompt.

We summarize the contributions as follows:
\begin{itemize}
\item We propose a novel heterogeneous graph interaction model with multi-granularity contrastive transfer learning (GIMC) to simultaneously address document-level and zero-shot cross-lingual event causality identification. 
\item We introduce a multi-granularity contrastive learning module to facilitate the cross-lingual transfer of language-agnostic causal knowledge, and construct a heterogeneous graph interaction network with four kinds of semantically rich nodes to model long-distance dependencies between events. 
\item Extensive experiments on the widely used multilingual ECI dataset show the effectiveness of our proposed model. F1 scores are improved by an average of 9.4\% and 8.2\% in monolingual and multilingual scenarios. 

\end{itemize}

\section{Methodology}
Figure \ref{fig4} shows the architecture of GIMC which consists of a heterogeneous graph interaction network (left) and a multi-granularity contrastive transfer learning module on graph (right). We first encode the document using multilingual pre-trained language model, then construct the heterogeneous graph interaction network with four types of nodes. Finally, we leverage statement-level and aspect-level casual pattern contrastive learning to facilitate the cross-lingual transfer of causal knowledge.

\subsection{Informative Phrase Extraction}
\label{sec3.1}
\begin{figure}[t]
	 	\centering{ 
	 	\includegraphics[width=0.49\textwidth]{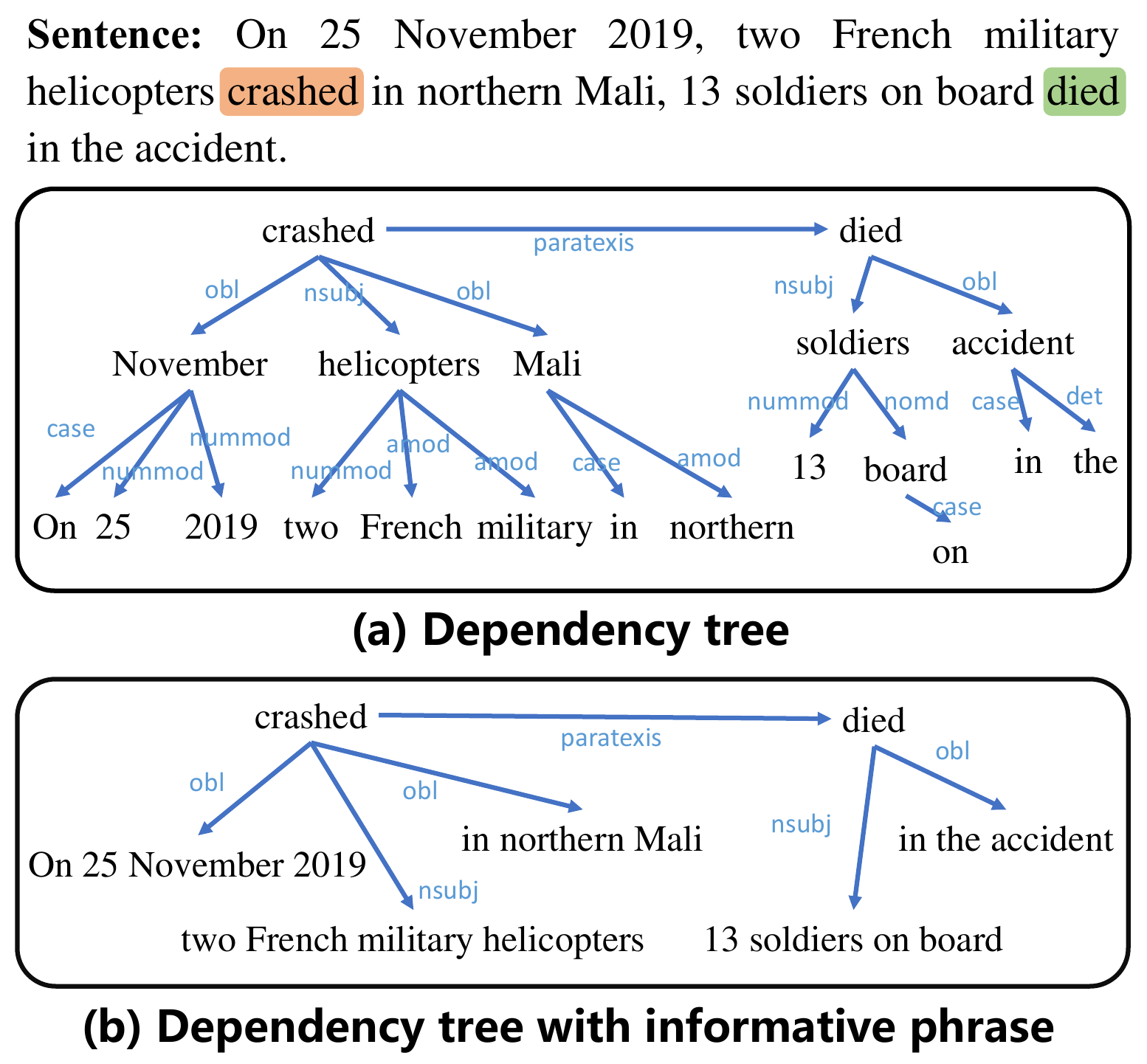}}  
	 	\caption{Example of informative phrase extraction}
	 	\label{fig3}
\end{figure}

Given a sentence $s = \{w_j\}^{|s|}_{j=1} \in \mathcal{D}$, we use the multilingual NLP toolkit Trankit \cite{nguyen-etal-2021-trankit}, which has an overall performance of about 93\% across different languages in sentence parsing, to obtain the dependency tree as shown in Figure \ref{fig3}(a). Previous studies only exploit the nodes in dependency tree as language-independent information to enhance ECI systems \cite{gao-etal-2019-modeling, tran-phu-nguyen-2021-graph} and overlook the rich semantics of dependency relations.

Thus, based on the semantics of the dependency relations \cite{de2008stanford, de2014universal, schuster2016enhanced}, we further process the dependency structure with heuristic rules to obtain a simplified dependency tree with informative phrases, as shown in Figure \ref{fig3}(b). We first analyze all dependency relations and their subtypes\footnote{\url{https://universaldependencies.org/u/dep/index.html}}, retaining 19 semantically rich and indicative dependency relations, e.g., \textit{nsubj} (nominal subject), \textit{obj} (object), \textit{obl} (oblique nominal). In this way, we extract the informative phrases and the corresponding dependency relations that indicates the relevant arguments of the events of interest (e.g., the subject (nsubj) of the event "\textit{\rm{crashed}}" is "\textit{\rm{two French military helicopters}}"). The complete list of dependency relations is in Appendix \ref{appendix A}. 

\subsection{Heterogeneous Graph Interaction Network}
\label{sec3.2}
We build a heterogeneous graph interaction network $\mathcal{G}$ which contains informative phrase nodes, sentence nodes, statement nodes and event pair nodes. After inserting special tags "\textit{<t>}" and "\textit{</t>}" at the start and end of all events to mark the event positions, we transform each word $w_i \in \mathcal{D}$ into the embedding $\bm{x}_i$ using pre-trained language model. For each informative phrase node $p$ which needs to incorporate its role information $r_p$, we initialize its embedding by $\bm{h}_p^{(0)}=\mathrm{Mean}(\{\bm{x}_j\}_{j \in p}) + \mathrm{Emb}(r_p)$. For each sentence node $s$, we initialize its embedding $\bm{h}_s^{(0)}=\mathrm{Mean}(\{\bm{x}_j\}_{j \in s})$. We define statement as a sentence containing two events to adapt to cross-sentence cases, such as the sentence 1 where the event "\textit{\rm{crashed}}" and the event "\textit{\rm{died}}" are located, or a concatenation of two sentences to include event pairs that across sentences, such as event "\textit{\rm{crashed}}" in sentence 1 and event "\textit{\rm{killing}}" in sentence 5. We initialize statement node embedding $\bm{h}_{st}^{(0)}=\mathrm{Mean}(\{\bm{x}_j\}_{j \in st})$. For event pair ($i, j$) with their representation ($\bm{e}_i, \bm{e}_j$), following \citet{chen-etal-2022-ergo}, we initialize $\bm{v}^{(0)}_{i,j}=\bm{W}_v [\bm{e}_i|| \bm{e}_j]$
, $\bm{W}_v$ are trainable parameters and $||$ denotes concatenation. 

To capture the interactions among these nodes, we introduce six types of edges:

\noindent\textbf{Phrase-Phrase Edge (P-P)} \quad The P-P edges are derived from the dependency structure. Informative phrase nodes are connected to each other by the dependency relations.

\noindent\textbf{Sentence-Phrase Edge (S-P)} \quad The phrase node is connected to its sentence node. The S-P edges enable to model the local contextual information of a phrase in its corresponding sentence.

\begin{figure*}[htbp]
	 	\centering  
	 	\includegraphics[width=1\textwidth]{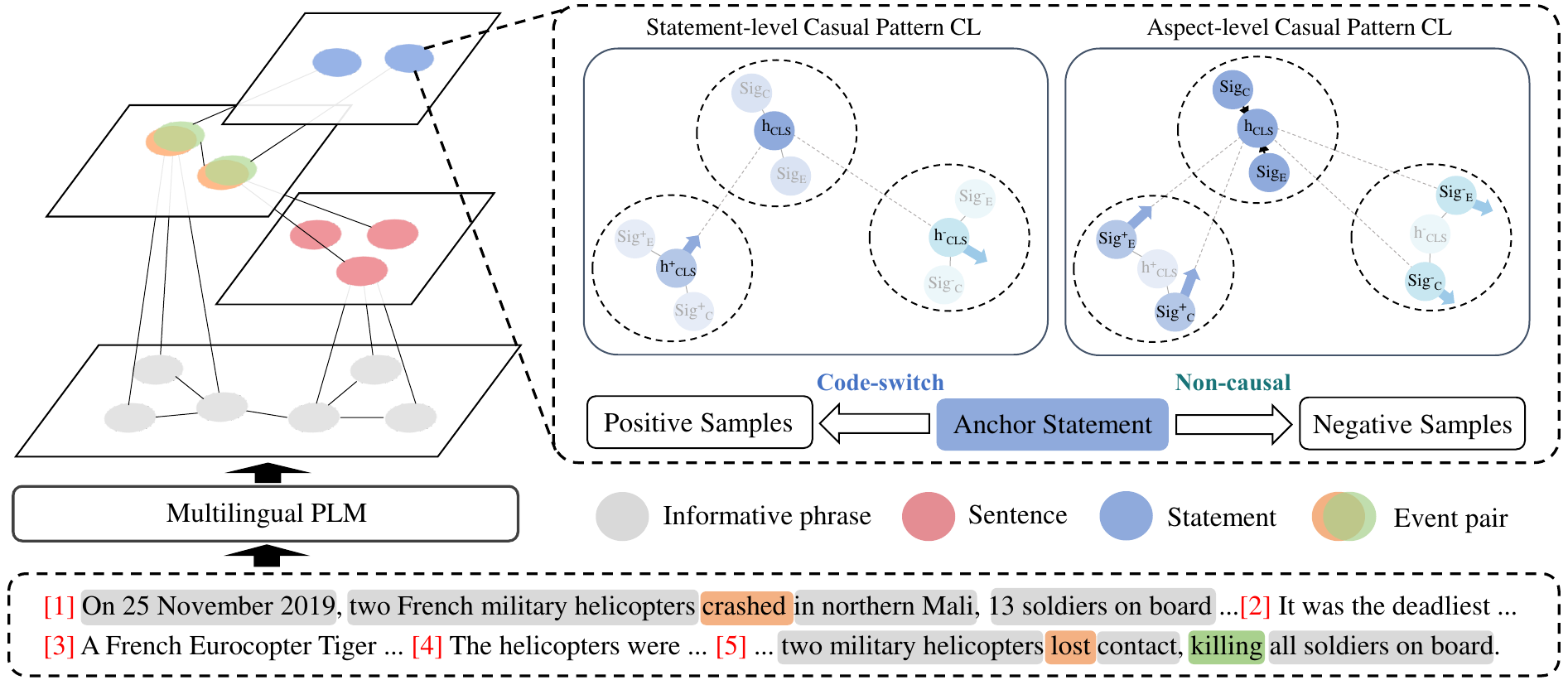}    
	 	\caption{Overview of our proposed heterogeneous graph interaction network with multi-granularity contrastive transfer learning (GIMC) for zero-shot cross-lingual document-level ECI.}
	 	\label{fig4}
\end{figure*}

\noindent\textbf{Phrase-Events Edge (P-E)} \quad The informative phrase containing an event mention can be seen as a more complete expression of the event. We model this information via the P-E edge.

\noindent\textbf{Sentence-Events Edge (S-E)} \quad If there are sentences containing any events of the current event pair, the event pair node is connected to the nodes of those sentences. We model the local context information of events with S-E edges.

\noindent\textbf{Statement-Events Edge (St-E)} \quad We connect the event node and its statement node. The St-E edge is expected to model the process of transfer the causality in statement to event pair to avoid the model from overfitting the causality in event pair.

\noindent\textbf{Events-Events Edge (E-E)} \quad There is an E-E edge between two event pair nodes only if the two corresponding event pairs share at least one event. E-E edge can promote the effective transmission of the causal information in event pairs.

After construct heterogeneous graph, we apply Graph Attention Networks v2 (GATv2) \cite{brody2022how} to model the global interactions. For each edge ($j,i$), the scoring function $f:\mathbb{R}^d \times \mathbb{R}^d \to \mathbb{R}$ indicates the importance of the features of the neighbor $j \in \mathcal{N}_i$ to the node $i$:

\begin{small}
\begin{equation} f(\bm{h}_i,\bm{h}_j)=\bm{a^{\top}}\mathrm{LeakyReLU}(\bm{W}\cdot[\bm{h}_i||\bm{h}_j])
\end{equation}
\end{small}

\noindent where $\bm{W}$ is denoted as $[\bm{W}_l||\bm{W}_r], \bm{W}_l\in\mathbb{R}^{d^\prime \times d}, \bm{W}_r\in\mathbb{R}^{d^\prime \times d}, \bm{a} \in \mathbb{R}^{d^\prime}$ are trainable parameters. $\bm{h}_i$ and $\bm{h}_j$ are the representations of node $i$ and $j$ respectively. Then the attention function is defined:

\begin{small}
\begin{equation}
		\alpha_{ij}=\frac{\mathrm{exp}(f(\bm{h}_i,\bm{h}_j))}{\sum_{j^\prime \in \mathcal{N}_i} \mathrm{exp}(f(\bm{h}_i,\bm{h}_j^\prime))}
\end{equation}
\end{small}

Then, we compute a weighted average of the transformed features of the neighbor nodes as the new representation of node $i$, using the normalized attention scores:

\begin{small}
\begin{equation}
		\bm{h}_i^\prime=\sum_{j \in \mathcal{N}_i}\alpha_{ij}\bm{W}_r\bm{h}_j
\end{equation}
\end{small}

We employ $K$ separate attention heads and concatenate their outputs as the output of node $i$:

\begin{small}
\begin{equation}
		\bm{h}_i^\prime=\bm{W}_o\big( \bigg|\bigg|_{k=1}^K\bm{h}_i^{\prime k}\big)
\end{equation}
\end{small}

\noindent where $\bm{W}_o$ are trainable parameters. 

\subsection{Multi-granularity Contrastive Learning Module}
\label{sec3.3}
Inspired by sentence-level ECI studies \cite{ijcai2020p499, zuo-etal-2021-improving}, which leverage the context-specific causal patterns of statements, for example, we can leverage the causal pattern “\textit{\rm{The [EVENT] \textbf{generates} [EVENT] ...}}” to identify the causation between \textit{\rm{\textbf{traffic congestion}}} and \textit{\rm{\textbf{environmental pollution}}} in a new statement “\textit{\rm The traffic congestion generates environmental pollution and economic loss}”. We generalize the causal pattern to the multilingual space, using contrastive learning to explicitly align causal representations across language. As shown in Figure \ref{fig4}, our contrastive transfer learning module consists of two part: 1) \textit{statement-level} causal pattern contrastive learning to align statement causal representation across languages, 2) \textit{aspect-level} causal pattern contrastive learning to align the representation between fine-grained causal patterns and statement. We select appropriate positive and negative samples for anchor statement:

\noindent\textbf{Positive Samples} \quad Given each anchor causal statement, we follow \citet{ijcai2020p533} to use MUSE bilingual dictionaries to generate multilingual code-switched statements as the positive samples. Unlike \citet{ijcai2020p533} which randomly replace a word at a time, we operate on phrases in statements. Specifically, given a phrase, we first randomly select a bilingual dictionary (e.g., en-da). Then, we switch each word in the phrase one by one using that bilingual dictionary, which is intuitively expected to maintain more complete semantic information. Taking the informative phrase in Figure \ref{fig3} as example, "\textit{\rm{two French military helicopters}}" $\xrightarrow{da}$ "\textit{\rm{to franske militær helikoptere}}".

\noindent\textbf{Negative Samples} \quad To make the negative examples more discriminative, we select non-causal statements within the same document as negative examples that do not have any text overlap with anchor statement. Furthermore, as we expect the model to focus on transfering across languages, we generate multilingual negative examples by code-switching to augment the list of negative samples. 

\noindent\textbf{1) \textit{Statement-level} causal pattern contrastive learning} \quad As \citet{zuo-etal-2021-improving} learn context-specific causal patterns from causal statements, we propose a statement-level causal pattern contrastive learning loss to explicitly align causal representations of anchor statement with the generated positive sample. Formally, this is formulated as:

\begin{scriptsize}
$$
		\mathcal{L}_{\mathrm{S}}=-\sum_{j=1}^nlog\frac{s(\bm{h}_{\mathrm{CLS}}, \bm{h}^j_{\mathrm{CLS}^+})}{s(\bm{h}_{\mathrm{CLS}}, \bm{h}^j_{\mathrm{CLS}^+})+\sum_{k=0}^{K-1}s(\bm{h}_{\mathrm{CLS}}, \bm{h}_{\mathrm{CLS}^-}^k)}
$$
\end{scriptsize}

\noindent where $s(\cdot)$ denotes dot product, $n$ is the number of corresponding positive samples, $K$ is the number of negative samples. $\bm{h}_{\mathrm{CLS}}$ is the embedding of CLS token, which serves as the initial state of statement and contains context-specific causal patterns. 

\noindent\textbf{2) \textit{Aspect-level} causal pattern contrastive learning} \quad Many previous studies \cite{cao-etal-2021-knowledge,chen-etal-2022-ergo} focus on exploring the causality of event pairs. Recently, \citet{ijcai2020p499} exploits event-agnostic, context-specific patterns which achieve promising performance. Thus, we categorize causal patterns into two aspects: event pair and event-agnostic context. Using the mentioned statement “\textit{\rm{The traffic congestion generates environmental pollution and economic loss}}” as an example, one of its causal patterns is the event pair aspect that "\textit{\rm{\textbf{traffic congestion}}}" and "\textit{\rm{\textbf{environmental pollution}}}", another causal pattern is the event-agnostic context aspect “\textit{\rm{The [EVENT] \textbf{generates} [EVENT] ...}}” Specifically, we split each statement into event pair and event-masked context, and get representations of these two types of aspects after encoding by multilingual pre-trained language model. 
We further introduce the \textit{aspect-level} causal pattern contrastive learning loss:

\begin{scriptsize}
$$
\mathcal{L}_{\mathrm{AspE}}=-\sum_{j=1}^nlog\frac{s(\bm{h}_{\mathrm{CLS}}, \bm{h}^j_{\mathrm{AspE^+}})}{s(\bm{h}_{\mathrm{CLS}}, \bm{h}^j_{\mathrm{AspE^+}})+\sum_{k=0}^{K-1}s(\bm{h}_{\mathrm{CLS}}, \bm{h}_{\mathrm{AspE}^-}^k)}
$$
\end{scriptsize}

\begin{scriptsize}
$$
\mathcal{L}_{\mathrm{AspC}}=-\sum_{j=1}^nlog\frac{s(\bm{h}_{\mathrm{CLS}}, \bm{h}^j_{\mathrm{AspC^+}})}{s(\bm{h}_{\mathrm{CLS}}, \bm{h}^j_{\mathrm{AspC^+}})+\sum_{k=0}^{K-1}s(\bm{h}_{\mathrm{CLS}}, \bm{h}_{\mathrm{AspC}^-}^k)}
$$
\end{scriptsize}

\noindent We consider contrastive learning loss from both event pair aspect ($\mathcal{L}_{_\mathrm{AspE}}$) and event-agnostic context aspect ($\mathcal{L}_{_\mathrm{AspC}}$).

\subsection{Training}
We concatenate event pair node representation $\bm{v}_{e_{i,j}}$, statement node representation $\bm{h}_{\mathrm{CLS}}$. For training, we adopt cross entropy as loss function:

\begin{small}
\begin{equation}
		\bm{p}_{e_{i},e_{j}}=\mathrm{softmax}(\bm{W}_p[\bm{v}_{e_{i,j}}||\bm{h}_{\mathrm{CLS}}])
\end{equation}
\end{small}
\begin{small}
\begin{equation}
		\mathcal{L}_{\mathrm{C}}=-\sum_{\bm{e}_{i},e_{j}\in E_s}\bm{y}_{e_{i},e_{j}}\mathrm{log}(\bm{p}_{e_{i},e_{j}})
\end{equation}
\end{small}

\noindent where $\bm{p}_{e_{i},e_{j}}$ is the predicted probability of causality between events $e_i$ and $e_j$. $E_s$ is the set of events, $\bm{y}_{e_{i},e_{j}}$ is a one-hot vector representing the gold label between $e_i$ and $e_j$. We sum the losses as follows:

\begin{small}
\begin{equation}
		\mathcal{L}_{\mathrm{all}}=\mathcal{L}_{_\mathrm{C}} + \mathcal{L}_{_\mathrm{S}} + \mathcal{L}_{_\mathrm{AspE}} + \mathcal{L}_{_\mathrm{AspC}}
\end{equation}
\end{small}
In our implementations, we employ the base versions of the language-specific pre-trained language models (PLMs) and the multilingual PLMs. The learning rate is initialized as 1e-3 with a linear decay. We use the AdamW algorithm \cite{loshchilov2017decoupled} to optimize model parameters. The batch size is set to 1, the number of GATv2 layers is 3. The number of training epochs is 60. Each experiment is conducted on NVIDIA GeForce RTX 3090 GPUs.

\section{Experiments}

\subsection{Dataset}
As the cross-lingual DECI is under-explored, the dataset we build upon is currently the only large-scale multi-lingual ECI dataset \citetlanguageresource{lai-etal-2022-meci} that employs a consistent annotation standard. It comprises as many as 3591 documents of five typologically diverse languages, i.e., English, Danish, Spanish, Turkish, and Urdu, the details are shown in Table \ref{tab_stat}. This dataset is not only larger but also more challenging, as a majority of events are 10 to 50 words away from each other in documents and there are clear divergences between the distance distributions of causal events over languages (we list the majority distance of causal events below).

\begin{table}[!ht]
\centering
\scriptsize
\resizebox{7.5cm}{!}{
\begin{tabular}{l|cccc}
\midrule
\textbf{Language} & \textbf{Document} & \textbf{Relation} & \textbf{Event} & \textbf{Distance}\\
\midrule
Danish & 519 & 1377 & 6909 &22\\
English &  438 & 2050 & 8732 &40\\
Spanish &  746 & 1312 & 11839 &39\\ 
Turkish &  1357 & 5337 & 14179 &33\\ 
Urdu &  531 & 979 & 4975 &23\\
\midrule
\end{tabular}}
\caption{Statistics for the MECI dataset.}
\label{tab_stat}
\end{table}

\begin{table*}
\centering
\resizebox{\linewidth}{!}{
\begin{tabular}{cc|ccc|ccc|ccc|ccc|ccc|c}
\toprule
\multicolumn{1}{c}{\multirow{2}{*}{}} &\multicolumn{1}{l}{\multirow{2}{*}{\textbf{Model}}}\vline & \multicolumn{3}{c}{\textbf{English}}\vline & \multicolumn{3}{c}{\textbf{Danish}}\vline & \multicolumn{3}{c}{\textbf{Spanish}}\vline & \multicolumn{3}{c}{\textbf{Turkish}}\vline & \multicolumn{3}{c}{\textbf{Urdu}}\vline & \multicolumn{1}{l}{\multirow{2}{*}{\textbf{AVG}}} \\

\cmidrule(lr){3-17}

\multicolumn{1}{c}{} & \multicolumn{1}{c}{}\vline & \textbf{P} & \textbf{R} & \textbf{F} & \textbf{P} & \textbf{R} & \textbf{F} & \textbf{P} & \textbf{R} & \textbf{F} & \textbf{P} & \textbf{R} & \textbf{F} & \textbf{P} & \textbf{R} & \textbf{F} & \multicolumn{1}{c}{}\\
\midrule

\multicolumn{1}{c}{\multirow{4}{*}{\rotatebox{90}{*}}} & \multicolumn{1}{l}{\textbf{PLM}}\vline & 35.6 & 44.9 & 39.7 & 23.2 & 23.0 & 23.1 & 42.7 & 44.6 & 43.6 & 40.4 & 56.0 & 46.9 & 20.2 & 33.5 & 25.2 & 35.7  \\

\multicolumn{1}{c}{} &\multicolumn{1}{l}{\textbf{ERGO}}\vline & 54.7 & 65.7 & 55.7 & 36.4 & 21.3 & 26.9 & 62.3 & 44.0 & 51.6 & 61.3 & 50.2 & 55.2 & 37.4 & 35.3 & 36.3 & 45.1 \\
\multicolumn{1}{c}{} &\multicolumn{1}{l}{\textbf{RichGCN}}\vline & 48.1 & \textbf{69.5} & 56.8 & 27.1 & 35.0 & 30.6 & 59.8 & 48.2 & 53.4 & 54.7 & 62.0 & 58.1 & 31.1 & \textbf{47.9} & 37.7 & 47.3 \\
\cmidrule(lr){2-18}
\rowcolor{gray!20}
\multicolumn{1}{c}{} &\multicolumn{1}{l}{\textbf{GIMC (base)}}\vline & \textbf{64.9} & 57.0 & \textbf{60.7} & \textbf{53.9} & \textbf{45.3} & \textbf{49.2} & \textbf{78.5} & \textbf{55.2} & \textbf{64.8} & \textbf{74.1} & \textbf{62.3} & \textbf{67.7} & \textbf{48.3} & 42.7 & \textbf{45.3} & \textbf{57.5} \\
\midrule
\multicolumn{1}{c}{\multirow{4}{*}{\rotatebox{90}{XLMR}}} & \multicolumn{1}{l}{\textbf{PLM}}\vline & 48.7 & 59.9 & 53.7 & 35.9 & 36.2 & 36.0 & 50.6 & 49.1 & 49.9 & 44.0 & 59.4 & 50.5 & 40.4 & 43.2 & 41.8 & 46.4 \\
\multicolumn{1}{c}{} &\multicolumn{1}{l}{\textbf{Know}}\vline & 39.3 & 42.6 & 40.9 & 31.4 & 11.4 & 16.7 & 39.9 & 28.4 & 33.2 & 36.5 & 46.7 & 41.0 & 41.1 & 22.2 & 28.9 & 32.1 \\

\multicolumn{1}{c}{} &\multicolumn{1}{l}{\textbf{ERGO}}\vline & 55.0 & 57.5 & 56.2 & 39.3 & 28.1 & 32.8 & 44.5 & 42.4 & 43.9 & 54.7 & 51.5 & 53.1 & 49.6 & 35.8 & 41.6 &  45.5 \\
\multicolumn{1}{c}{} &\multicolumn{1}{l}{\textbf{RichGCN}}\vline & 50.6 & \textbf{68.0} & 58.1 & 31.9 & 50.0 & 38.9 & 50.7 & 55.0 & 52.8 & 50.5 & \textbf{64.6} & 56.7 & 37.7 & \textbf{56.0} & 45.1  & 50.3 \\
\cmidrule(lr){2-18}
\rowcolor{gray!20}
\multicolumn{1}{c}{} &\multicolumn{1}{l}{\textbf{GIMC (base)}}\vline & \textbf{61.5} & 58.4 & \textbf{59.9} & \textbf{52.1} & \textbf{50.5} & \textbf{51.3} & \textbf{81.1} & \textbf{56.4} & \textbf{66.5} & \textbf{68.7} & 58.2 & \textbf{63.0} & \textbf{64.3} & 42.6 & \textbf{51.2} & \textbf{58.4} \\
\midrule

\multicolumn{1}{c}{\multirow{4}{*}{\rotatebox{90}{mBERT}}} &\multicolumn{1}{l}{\textbf{PLM}}\vline & 38.4 & 46.0 & 41.9 & 25.2 & 26.6 & 25.9 & 43.9 & 41.5 & 42.7 & 36.2 & 48.7 & 41.6 & 31.9 & 34.3 & 33.0 & 37.0 \\
\multicolumn{1}{c}{} &\multicolumn{1}{l}{\textbf{Know}}\vline & 35.8 & 56.7 & 43.9 & 25.8 & 36.0 & 30.1 & 39.7 & 38.3 & 39.0 & 39.7 & 46.9 & 43.0 & 36.7 & 35.3 & 36.0 & 38.4 \\

\multicolumn{1}{c}{} &\multicolumn{1}{l}{\textbf{ERGO}}\vline & 58.2 & 49.0 & 53.2 & 34.4 & 24.6 & 28.7 & 56.3 & 39.7 & 46.6 & 52.7 &45.5 & 48.8 & 43.4 & 41.6  & 42.5 & 43.9 \\
\multicolumn{1}{c}{} &\multicolumn{1}{l}{\textbf{RichGCN}}\vline & 48.4 & \textbf{67.1} & 56.2 & 29.7 & 38.0 & 33.4 & 51.2 & 52.0 & 51.6 & 50.0 & 59.9 & 54.5 & 40.1 & \textbf{50.0} & 44.5  & 48.0 \\
\cmidrule(lr){2-18}
\rowcolor{gray!20}
\multicolumn{1}{c}{} &\multicolumn{1}{l}{\textbf{GIMC (base)}}\vline & \textbf{63.4} & 54.8 & \textbf{58.8} & \textbf{60.2} & \textbf{45.2} & \textbf{51.6} & \textbf{77.5} & \textbf{55.7} & \textbf{64.8} & \textbf{70.1} & \textbf{60.1} & \textbf{64.7} & \textbf{62.1} & 42.4 & \textbf{50.4} & \textbf{58.1} \\
\midrule
\multicolumn{2}{c}{\textbf{GPT-3.5 (zero-shot)}}\vline & 24.6 & 79.4 & 37.6 & 10.0 & 66.5 & 17.4 & 7.3 & 74.2 & 13.3 & 27.0 & 69.3 & 38.8 & 15.7 & 63.8 & 25.2 & 26.4 \\
\bottomrule
\end{tabular}}

\caption{\label{tab2} Monolingual performance on MECI dataset. We report the results using language-specific PLMs ("*"), XLMR and mBERT as the backbone respectively. AVG is the average F1 score for five languages.}
\end{table*}

\begin{table*}
\centering
\resizebox{\linewidth}{!}{
\begin{tabular}{cc|ccc|ccc|ccc|ccc|ccc|cc}
\toprule
\multicolumn{1}{c}{\multirow{2}{*}{}} &\multicolumn{1}{l}{\multirow{2}{*}{\textbf{Model}}}\vline & \multicolumn{3}{c}{\textbf{English} $\to$ \textbf{English}}\vline & \multicolumn{3}{c}{\textbf{English} $\to$ \textbf{Danish}}\vline & \multicolumn{3}{c}{\textbf{English} $\to$ \textbf{Spanish}}\vline & \multicolumn{3}{c}{\textbf{English} $\to$ \textbf{Turkish}}\vline & \multicolumn{3}{c}{\textbf{English} $\to$ \textbf{Urdu}}\vline & \multicolumn{1}{l}{\multirow{2}{*}{\textbf{AVG}}} & \multicolumn{1}{c}{\multirow{2}{*}{\textbf{$\Delta$}}} \\

\cmidrule(lr){3-17}

\multicolumn{1}{c}{} & \multicolumn{1}{c}{}\vline & \textbf{P} & \textbf{R} & \textbf{F} & \textbf{P} & \textbf{R} & \textbf{F} & \textbf{P} & \textbf{R} & \textbf{F} & \textbf{P} & \textbf{R} & \textbf{F} & \textbf{P} & \textbf{R} & \textbf{F} & \multicolumn{1}{c}{} & \multicolumn{1}{c}{}\\
\midrule

\multicolumn{1}{c}{\multirow{4}{*}{\rotatebox{90}{XLMR}}} & \multicolumn{1}{l}{\textbf{PLM}}\vline & 48.7 & 59.9 & 53.7 & 20.1 & \textbf{59.2} & 30.1 & 16.0 & \textbf{66.4} & 25.8 & 36.1 & \textbf{60.5} & 45.2 & 25.7 & \textbf{62.0} & 36.3 & 38.2 & 19.4\\
\multicolumn{1}{c}{} &\multicolumn{1}{l}{\textbf{Know}}\vline & 39.3 & 42.6 & 40.9 & 13.3 & 42.1 & 20.3 & 10.4 & 47.3 & 17.1 & 25.8 & 57.6 & 35.7 & 19.3 & 54.5 & 28.5 & 28.5 & 15.5\\

\multicolumn{1}{c}{} &\multicolumn{1}{l}{\textbf{ERGO}}\vline & 55.0 & 57.5 & 56.2 & 36.4 & 34.6 & 35.5 & 34.0 & 52.8 & 41.4 & 45.3 & 40.9 & 43.0 & 38.2 & 42.9 & 40.0  & 43.2 & 16.2\\
\multicolumn{1}{c}{} &\multicolumn{1}{l}{\textbf{RichGCN}}\vline & 50.6 & \textbf{68.0} & 58.1 & 28.5 & 43.7 & 34.5 & 22.7 & 62.4 & 33.3 & 46.4 & 55.0 & 50.3 & 38.6 & 55.2 & 45.5  & 44.3 & 17.2\\
\cmidrule(lr){2-19}
\rowcolor{gray!20}
\multicolumn{1}{c}{} &\multicolumn{1}{l}{\textbf{GIMC (Ours)}}\vline & \textbf{64.2} & 54.3 & \textbf{58.8} & \textbf{44.5} & 44.1 & \textbf{44.3} & \textbf{69.0} & 40.4 & \textbf{51.0} & \textbf{56.7} & 46.7 & \textbf{51.2} & \textbf{61.5} & 38.5 & \textbf{47.3} & \textbf{50.5} & \textbf{10.4}\\
\midrule

\multicolumn{1}{c}{\multirow{4}{*}{\rotatebox{90}{mBERT}}} &\multicolumn{1}{l}{\textbf{PLM}}\vline & 38.4 & 46.0 & 41.9 & 12.4 & 35.4 & 18.4 & 11.4 & 63.3 & 19.3 & 21.5 & 47.6 & 29.6 & 17.0 & 44.2 & 24.6 & 26.7 & 18.9\\
\multicolumn{1}{c}{} &\multicolumn{1}{l}{\textbf{Know}}\vline & 35.8 & 56.7 & 43.9 & 7.8 & \textbf{62.0} & 13.8 & 7.2 & \textbf{69.4} & 13.0 & 20.4 & \textbf{55.5} & 29.9 & 14.2 & \textbf{61.5} & 23.0 & 24.7 & 24.0\\

\multicolumn{1}{c}{} &\multicolumn{1}{l}{\textbf{ERGO}}\vline & 58.2 & 49.0 & 53.2 & 32.0 & 28.7 &30.3 & 35.4 & 41.3 & 38.1 & 46.3 & 33.0 & 38.6 & 35.3 & 36.4 & 35.8  & 39.2 & 17.5\\
\multicolumn{1}{c}{} &\multicolumn{1}{l}{\textbf{RichGCN}}\vline & 48.4 & \textbf{67.1} & 56.2 & 23.7 & 45.3 & 31.1 & 20.6 & 58.6 & 30.5 & 44.5 & 52.0 & 48.0 & 35.0 & 56.8 & 43.3  & 41.8 & 18.0\\
\cmidrule(lr){2-19}
\rowcolor{gray!20}
\multicolumn{1}{c}{} &\multicolumn{1}{l}{\textbf{GIMC (Ours)}}\vline & \textbf{66.6} & 54.0 & \textbf{59.6} & \textbf{56.3} & 41.9 & \textbf{48.0} & \textbf{58.2} & 50.2 & \textbf{53.9} & \textbf{61.6} & 42.5 & \textbf{50.3} & \textbf{55.7} & 43.4 & \textbf{48.8} & \textbf{52.1} & \textbf{9.4}\\
\midrule
\multicolumn{2}{c}{\textbf{GPT-3.5 (few-shot)}}\vline & 27.9 & 83.2 & 41.8 & 12.7 & 75.0 & 21.8 & 8.8 & 84.9 & 15.9 & 27.3 & 80.2 & 40.8 & 18.2 & 75.6 & 29.4 & 29.9 & 14.8\\
\bottomrule
\end{tabular}}

\caption{\label{tab3}Zero-shot cross-lingual document-level ECI performance with English as the source language. We report the average (AVG) and fluctuation ($\Delta$) of F1 score under the different multilingual PLMs.}
\end{table*}

\subsection{Experimental Settings}

We utilize the base versions of all PLMs and evaluate our model in the following settings: 

\noindent\textbf{Monolingual learning setting} \quad The training and test data are in same language. We use language specific PLMs, i.e., BotXO2\footnote{\url{https://huggingface.co/Maltehb/danish-bert-botxo}} for Danish, BERT \cite{devlin-etal-2019-bert} for English, BETO \cite{canete2020spanish} for Spanish, BERTurk \cite{stefan_schweter_2020_3770924} for Turkish, and UrduHack\footnote{\url{https://github.com/urduhack/urduhack}} for Urdu.


\noindent\textbf{Multilingual learning  setting} \quad The ECI models are trained in source language and directly tested in target language. We utilize the multilingual PLMs, i.e., mBERT \cite{devlin-etal-2019-bert} or XLMR \cite{conneau-etal-2020-unsupervised} as the backbone.

\subsection{Baselines}


We compare our method with the vanilla PLM, LLM and three strong monolingual baselines (we replace their backbones with multilingual PLM):

\noindent (1) \textbf{PLM} \quad By leveraging the embeddings $\bm{h}_i, \bm{h}_j$ of two events, the overall representation vector is formed $\bm{h}_{i,j}=[\bm{h}_i,\bm{h}_j,\bm{h}_i-\bm{h}_j,\bm{h}_i*\bm{h}_j]$ for ECI. 

\noindent (2) \textbf{Know} \quad \citet{ijcai2020p499} retrieves external knowledge from knowledge base and event-agnostic, context-specific patterns to enrich the representations of events for causality identification.

\begin{table*}
\centering
\resizebox{\linewidth}{!}{
\begin{tabular}{c|ccc|ccc|ccc|ccc|ccc|cc}
\toprule
\multicolumn{1}{l}{\multirow{2}{*}{\textbf{Model}}}\vline & \multicolumn{3}{c}{\textbf{Danish} $\to$ \textbf{English}}\vline & \multicolumn{3}{c}{\textbf{Danish} $\to$ \textbf{Danish}}\vline & \multicolumn{3}{c}{\textbf{Danish} $\to$ \textbf{Spanish}}\vline & \multicolumn{3}{c}{\textbf{Danish} $\to$ \textbf{Turkish}}\vline & \multicolumn{3}{c}{\textbf{Danish} $\to$ \textbf{Urdu}}\vline & \multicolumn{1}{l}{\multirow{2}{*}{\textbf{AVG}}} & \multicolumn{1}{c}{\multirow{2}{*}{\textbf{$\Delta$}}} \\

\cmidrule(lr){2-16}

\multicolumn{1}{c}{}\vline & \textbf{P} & \textbf{R} & \textbf{F} & \textbf{P} & \textbf{R} & \textbf{F} & \textbf{P} & \textbf{R} & \textbf{F} & \textbf{P} & \textbf{R} & \textbf{F} & \textbf{P} & \textbf{R} & \textbf{F} & \multicolumn{1}{c}{} & \multicolumn{1}{c}{}\\

\midrule

 \multicolumn{1}{l}{\textbf{PLM}}\vline & 43.6 & 25.9 & 32.5 & 32.1 & 25.6 & 28.5 & 28.7 & 35.7 & 31.8 & 29.6 & 22.5 & 25.6 & 28.6 & 20.3 & 23.7 & 28.4 & 3.8\\

\multicolumn{1}{l}{\textbf{ERGO}}\vline & 67.1 & 30.0 & 41.4 & 34.4 & 24.6 & 28.7 & 57.3 & 33.1 & 41.9 & 52.5 & 31.0 & 38.9 & 39.2 & 28.5  & 33.0 & 36.7 & 10.1\\
\multicolumn{1}{l}{\textbf{RichGCN}}\vline & 52.8 & 40.0 & 45.3 & 29.7 & 29.0 & 33.4 & 34.4 & \textbf{49.3} & 40.5 & 44.4 & \textbf{42.3} & 43.3 & 32.4 & \textbf{54.7}  & 40.6 & 40.7 & 9.0\\
\rowcolor{gray!20}
\multicolumn{1}{l}{\textbf{GIMC (Ours)}}\vline & \textbf{63.7} & \textbf{42.7} & \textbf{51.1} & \textbf{61.1} & \textbf{40.4} & \textbf{48.6} & \textbf{65.8} & 44.9 & \textbf{53.4} & \textbf{63.3} & 37.1 & \textbf{46.8} & \textbf{57.6} & 41.6 & \textbf{48.3} & \textbf{49.6} & \textbf{2.4}\\
\midrule
\multicolumn{1}{l}{\textbf{GPT-3.5 (few-shot)}}\vline & 23.2 & 80.8 & 36.1 & 10.4 & 79.5 & 18.4 & 7.9 & 85.6 & 14.4 & 25.2 & 79.7 & 38.3 & 17.2 & 82.7 & 28.5 & 27.1 & 12.3\\

\midrule
\midrule

\multicolumn{1}{l}{\textbf{Model}}\vline & \multicolumn{3}{c}{\textbf{Spanish} $\to$ \textbf{English}}\vline & \multicolumn{3}{c}{\textbf{Spanish} $\to$ \textbf{Danish}}\vline & \multicolumn{3}{c}{\textbf{Spanish} $\to$ \textbf{Spanish}}\vline & \multicolumn{3}{c}{\textbf{Spanish} $\to$ \textbf{Turkish}}\vline & \multicolumn{3}{c}{\textbf{Spanish} $\to$ \textbf{Urdu}}\vline & \multicolumn{1}{l}{\textbf{AVG}} & \multicolumn{1}{c}{\textbf{$\Delta$}} \\

\midrule

\multicolumn{1}{l}{\textbf{PLM}}\vline & 57.1 & 23.2 & 33.0 & 27.8 & 16.3 & 20.6 & 51.3 & 45.5 & 48.2 & 52.9 & 18.5 & 27.4 & 35.0 & 21.7 & 26.8 & 31.2 & 21.3 \\

\multicolumn{1}{l}{\textbf{ERGO}}\vline & 71.0 & 21.7 & 33.3 & 42.5 & 20.1 & 27.2 & 56.3 & 39.7 & 46.6 & 55.3 & 21.4 & 30.8 & 44.9 & 24.6  & 31.7 & 33.9 & 15.8\\
\multicolumn{1}{l}{\textbf{RichGCN}}\vline & 56.4 & 31.1 & 40.2 & 26.9 & 27.7 & 27.3 & 51.2 & \textbf{52.0} & 51.6 & 52.5 & 29.9 & 38.1 & 42.1 & 33.7  & 37.5 & 38.9 & \textbf{15.8}\\
\rowcolor{gray!20}
\multicolumn{1}{l}{\textbf{GIMC (Ours)}}\vline & \textbf{75.6} & \textbf{36.7} & \textbf{49.4} & \textbf{65.5} & \textbf{35.8} & \textbf{46.3} & \textbf{81.4} & 51.3 & \textbf{62.9} & \textbf{70.0} & \textbf{33.4} & \textbf{45.2} & \textbf{66.1} & \textbf{36.6} & \textbf{47.1} & \textbf{50.2} & 15.9\\
\midrule
\multicolumn{1}{l}{\textbf{GPT-3.5 (few-shot)}}\vline & 23.7 & 85.7 & 37.2 & 9.9 & 77.5 & 17.5 & 6.9 & 84.0 & 12.8 & 23.9 & 80.0 & 36.9 & 15.5 & 80.6 & 26.0 & 26.1 & 16.6\\
\midrule
\midrule

\multicolumn{1}{l}{\textbf{Model}}\vline & \multicolumn{3}{c}{\textbf{Turkish} $\to$ \textbf{English}}\vline & \multicolumn{3}{c}{\textbf{Turkish} $\to$ \textbf{Danish}}\vline & \multicolumn{3}{c}{\textbf{Turkish} $\to$ \textbf{Spanish}}\vline & \multicolumn{3}{c}{\textbf{Turkish} $\to$ \textbf{Turkish}}\vline & \multicolumn{3}{c}{\textbf{Turkish} $\to$ \textbf{Urdu}}\vline & \multicolumn{1}{l}{\textbf{AVG}} & \multicolumn{1}{c}{\textbf{$\Delta$}} \\

\midrule

\multicolumn{1}{l}{\textbf{PLM}}\vline & 34.3 & 43.3 & 38.3 & 24.5 & 33.5 & 28.3 & 23.4 & 45.4 & 30.9 & 39.8 & 50.9 & 44.7 & 27.2 & 43.5 & 33.5 & 35.1 & 11.9\\

\multicolumn{1}{l}{\textbf{ERGO}}\vline & 49.6 & 32.6 & 39.3 & 27.6 & 28.1 & 27.8 &29.1 & 40.8 & 33.9 & 52.7 &45.5 & 48.8 &38.0 & 45.8  & 41.5 & 38.2 & 13.1\\
\multicolumn{1}{l}{\textbf{RichGCN}}\vline & 44.5 & \textbf{61.3} & 51.6 & 20.3 & \textbf{48.4} & 28.6 & 22.6 & \textbf{61.3} & 33.1 & 50.0 & \textbf{59.9} & 54.5 & 36.4 & \textbf{59.2}  & 45.1 & 42.5 & 14.9\\
\rowcolor{gray!20}
\multicolumn{1}{l}{\textbf{GIMC (Ours)}}\vline & \textbf{58.8} & 47.7 & \textbf{52.7} & \textbf{64.8} & 39.6 & \textbf{49.2} & \textbf{61.8} & 55.8 & \textbf{58.6} & \textbf{76.2} & 56.2 & \textbf{64.7} & \textbf{62.3} & 54.2 & \textbf{58.0} & \textbf{56.6} & \textbf{10.1}\\
\midrule
\multicolumn{1}{l}{\textbf{GPT-3.5 (few-shot)}}\vline & 17.1 & 82.1 & 28.3 & 5.8 & 70.2 & 10.8 & 4.0 & 76.4 & 7.7 & 17.7 & 78.4 & 28.9 & 32.2 & 100 & 48.7 & 24.9 & 14.9\\
\midrule
\midrule

\multicolumn{1}{l}{\textbf{Model}}\vline & \multicolumn{3}{c}{\textbf{Urdu} $\to$ \textbf{English}}\vline & \multicolumn{3}{c}{\textbf{Urdu} $\to$ \textbf{Danish}}\vline & \multicolumn{3}{c}{\textbf{Urdu} $\to$ \textbf{Spanish}}\vline & \multicolumn{3}{c}{\textbf{Urdu} $\to$ \textbf{Turkish}}\vline & \multicolumn{3}{c}{\textbf{Urdu} $\to$ \textbf{Urdu}}\vline & \multicolumn{1}{l}{\textbf{AVG}} & \multicolumn{1}{c}{\textbf{$\Delta$}} \\

\midrule

\multicolumn{1}{l}{\textbf{PLM}}\vline & 38.8 & 25.5 & 30.8 & 18.7 & 21.1 & 19.8 & 30.4 & 23.7 & 26.6 & 29.5 & 21.4 & 24.8 & 45.1 & 33.6 & 38.5 & 28.1 & 13.0\\

\multicolumn{1}{l}{\textbf{ERGO}}\vline & 52.7 & 25.0 & 33.9 & 22.2 & 22.8 & 22.5 & 47.5 &27.5 & 34.8 & 51.3 & 33.0 & 40.1 & 43.4 & 41.6  & 42.5 & 34.7 & 9.7\\
\multicolumn{1}{l}{\textbf{RichGCN}}\vline & \textbf{54.5} & 34.8 & 42.5 & 26.2 & \textbf{29.3} & 27.7 & 34.7 & 41.0 & 37.6 & 56.4 & \textbf{32.6} & \textbf{41.3} & 40.1 & \textbf{50.0}  & 44.5 & 38.7 & \textbf{7.2}\\
\rowcolor{gray!20}
\multicolumn{1}{l}{\textbf{GIMC (Ours)}}\vline & 52.5 & \textbf{38.3} & \textbf{44.3} & \textbf{50.5} & 28.7 & \textbf{36.6} & \textbf{51.3} & \textbf{46.3} & \textbf{48.7} & \textbf{61.1} & 24.7 & 35.2 & \textbf{60.5} & 43.8 & \textbf{50.8} & \textbf{43.1} & 9.6\\
\midrule
\multicolumn{1}{l}{\textbf{GPT-3.5 (few-shot)}}\vline & 20.7 & 89.0 & 33.7 & 8.3 & 84.6 & 15.1 & 5.6 & 87.7 & 10.6 & 21.8 & 87.1 & 35.0 & 8.3 & 84.6 & 15.1 & 21.9 & 10.7\\
\bottomrule
\end{tabular}
}
\caption{\label{tab4} The Performance in low-resource settings. We separately report the cross-lingual performance with low-source language Danish, Spanish, Turkish, and Urdu as source language.}
\end{table*}

\noindent (3) \textbf{RichGCN} \quad \citet{tran-phu-nguyen-2021-graph} implement several interaction graphs, by learning a linear combination of the adjacency matrices of these graphs to get a final graph and uses graph convolutional network (GCN) to capture capture relevant context information.

\noindent (4) \textbf{ERGO} \quad \citet{chen-etal-2022-ergo} define a pair of events as a node and build a complete event relational graph to capture the causation transitivity among event pairs via a graph transformer.

\noindent (5) \textbf{GPT-3.5} \quad We use gpt-3.5-turbo with zero-shot / few-shot learning on this complex multilingual task. The prompt is shown in Figure \ref{fig_prompt} and language-specific prompts can be found in Appendix \ref{appendix B}. 

\begin{figure}[t]
	 	\centering{ 
	 	\includegraphics[width=0.49\textwidth]{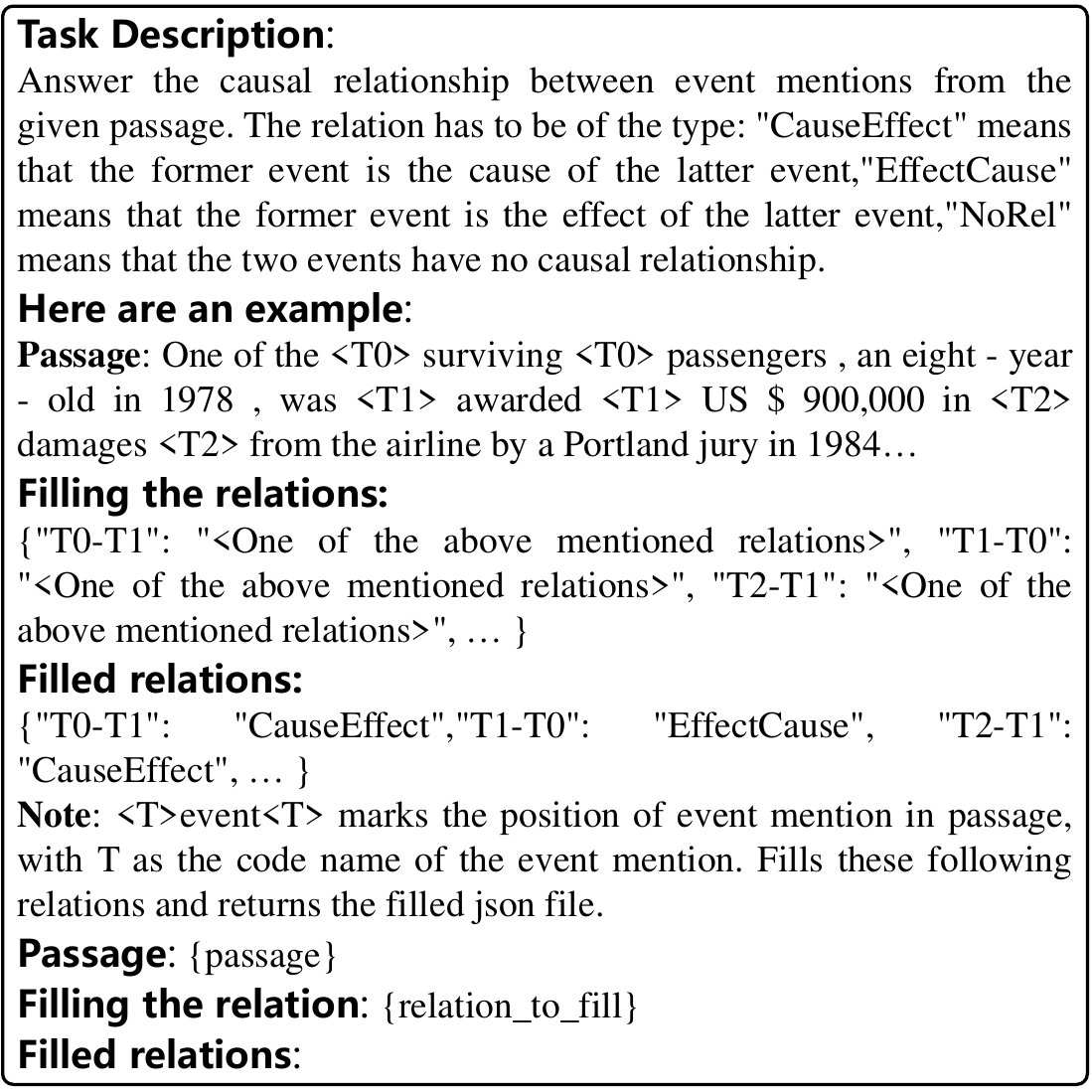}}  
	 	\caption{Few-shot prompt for English. We set language specific prompts for different source languages. The passage and relations are truncated. }
            \vspace{-0.5cm} 
	 	\label{fig_prompt}
\end{figure}

\subsection{Main Results}
\noindent\textbf{(1) The Performance in Monolingual Setting} \quad 
The results is illustrated in Table \ref{tab2}, in this monolingual setting, we observe that our model GIMC (base version), which removes contrastive learning module, achieves the best performance across all languages, and outperforms RichGCN across all settings with an average improvement of 9.4\% in F1 score. It is worth noting that we test the effectiveness of our method with different PLMs as the backbone. The significant improvement of our method over the vanilla PLM (18.8\%) suggests that the success of our model can be attributed to the effectiveness of our method rather than the backbone. These baselines without well-designed cross-lingual module, rely almost on multilingual backbones, perform poorly when transferring to other languages. However our model can cope well with all language settings, even for Danish (an average increase of 16.4\% compared with RichGCN).

\noindent\textbf{(2) The Performance in High-Resource Setting} \quad 
As shown in Table \ref{tab3}, the entire GIMC performs surprisingly well on all languages. Comparing with the results of the baseline RichGCN with mBERT, GIMC pushes the average F1 score from 41.8\% to 52.1\% and effectively reduces performance fluctuations (i.e., $\Delta$) during transfer to other languages. Note that the performance of RichGCN with mBERT drops significantly for three target languages Danish (by 25.1\%), Spanish (by 25.7\%), and Urdu (by 12.9\%), we observed the same trend in our experiments with GPT-3.5, demonstrating that despite extensive training on high-resource language English, LLM fails to achieve plug-and-play functionality when faced with complex cross-lingual tasks, which is consistent with \citet{lai2023chatgpt}.

\begin{table}
\centering
\renewcommand\arraystretch{1.25}
\resizebox{\linewidth}{!}{
\begin{tabular}{c|ccccc|cc}
\toprule

\multicolumn{1}{l}{\textbf{Model}}\vline & \textbf{English} & \textbf{Danish} & \textbf{Spanish} & \textbf{Turkish} & \textbf{Urdu} & \textbf{AVG} & \textbf{$\Delta$} \\

\midrule

\multicolumn{1}{l}{\textbf{GIMC}}\vline & \textbf{59.6} & 48.0 & \textbf{53.9} & \textbf{50.3} & \textbf{48.8} & \textbf{52.1} & \textbf{9.4} \\

\multicolumn{1}{l}{\textbf{GIMC(GCN)}}\vline & 58.6 & 47.5 & 50.4 & 47.1 & 48.3 & 50.4 & 10.2 \\

\multicolumn{1}{c}{\textbf{- P-P}}\vline & 57.3 & 43.2 & 49.2 & 46.6 & 45.4 & 48.3 & 11.2 \\

\multicolumn{1}{c}{\textbf{- S-P}}\vline & 58.3 & 46.4 & 53.0 & 49.2 & 46.3 & 50.6 & 9.6 \\

\multicolumn{1}{c}{\textbf{- P-E}}\vline & 58.8 & 47.5 & 52.2 & 48.3 & 47.2 & 50.8 & 10.0\\

\multicolumn{1}{c}{\textbf{- S-E}}\vline & 57.9 & 45.5 & 51.5 & 46.1 & 47.6 & 49.7 & 10.2 \\

\multicolumn{1}{c}{\textbf{- St-E}}\vline & 58.9 & \textbf{48.4} & 51.2 & 47.5 & 46.1 & 50.4 & 10.6 \\

\multicolumn{1}{c}{\textbf{- E-E}}\vline & 57.5 & 43.3 & 47.6 & 45.7 & 45.5 & 47.9 & 12.0 \\

\bottomrule
\end{tabular}
}

\caption{\label{tab5}The ablation experiments of GIMC’s heterogeneous graph interaction network. We directly report the F1 scores (\%) in the zero-shot cross-lingual setting using English as source language.}
\end{table}

\begin{table}
\centering
\renewcommand\arraystretch{1.25}
\resizebox{\linewidth}{!}{
\begin{tabular}{c|ccccc|cc}
\toprule

\multicolumn{1}{l}{\textbf{Model}}\vline & \textbf{English} & \textbf{Danish} & \textbf{Spanish} & \textbf{Turkish} & \textbf{Urdu} & \textbf{AVG} & \textbf{$\Delta$} \\

\midrule

\multicolumn{1}{l}{\textbf{GIMC}}\vline & 59.6 & \textbf{48.0} & \textbf{53.9} & 50.3 & \textbf{48.8} & \textbf{52.1} & \textbf{9.4} \\

\multicolumn{1}{l}{\textbf{GIMC$_{word}$}}\vline & \textbf{61.2} & 47.2 & 51.6 & \textbf{50.6} & 48.3 & 51.8 & 11.8 \\

\multicolumn{1}{l}{\textbf{GIMC(base)}}\vline & 58.8 & 44.2 & 49.7 & 46.8 & 47.6 & 49.4 & 11.7 \\

\bottomrule
\end{tabular}
}

\caption{\label{tab6}Performance of GIMC on ablation study for  cross-lingual contrastive transfer learning module.}
\end{table}

\begin{figure}
   \centering{ 
   \includegraphics[width=0.48\textwidth]{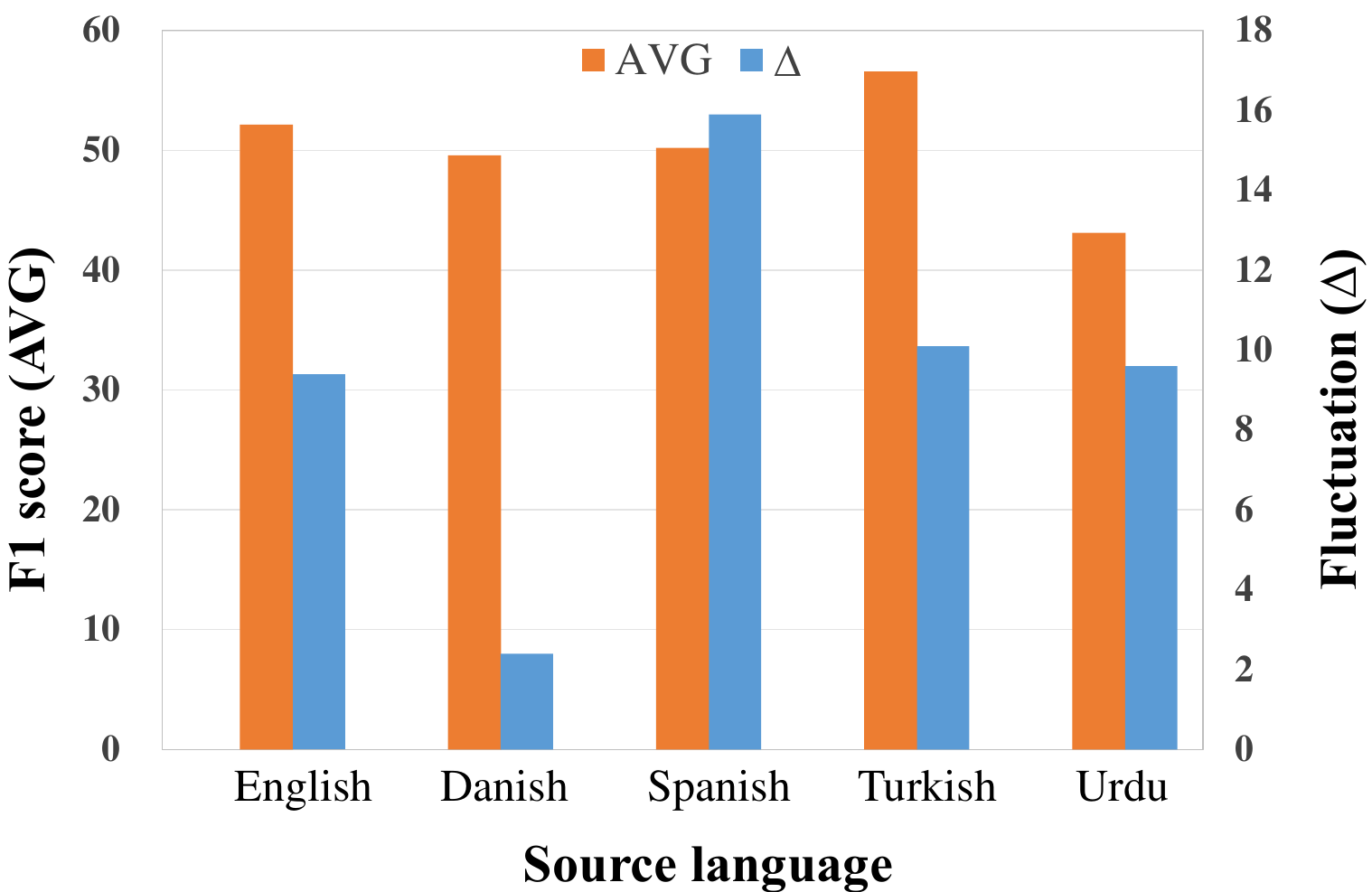}}  
   \caption{Average F1 score (AVG) and fluctuation ($\Delta$) of GIMC with different source language.}
   \label{fig5}
\end{figure}

\noindent\textbf{(3) The Performance in Low-Resource Setting} \quad Since English as high-resource language may benefit more from PLMs than other languages, we further conduct experiments with low-resource language as source language. We utilize mBERT as our backbone due to its superior performance in the above experiments. As shown in Table \ref{tab4}, GIMC improves average performance and fluctuation in all language settings and achieves double first in monolingual performance (by 64.7\%) and cross-lingual transfer performance (by 56.6\%) on Turkish. Contrarily, the GPT-3.5 performs unsatisfactorily (with an average performance 24.9\% lower than GIMC). Moreover, The drop of RichGCN for Urdu → Turkish is smaller compared to ours. It suggests that the baseline might be sensitive to certain specific languages, whereas our model demonstrates consistently strong performance across all settings. 

\subsection{Effects of Heterogeneous Graph Interaction Network}
To investigate the effectiveness of our graph interaction network, we conduct the ablation studies as shown in Table \ref{tab5}, the F1 score would decreases 1.3\% \textasciitilde 3.8\% without the top five types of edges. Besides, removing the E-E edge causes an significant drop by 4.2\%, which is consistent with "preserving transitivity of causation" \cite{paul2013causation, chen-etal-2022-ergo}, each edge represents a possible causal pattern transitivity between the event pair. Despite a slight decrease in performance when using other information aggregation methods such as GCN, our method is still competitive. Ablation experiments demonstrates that our graph interaction network helps improve the performance.

\subsection{Effects of Cross-lingual Contrastive Transfer Learning Module}

As depicted in the Table \ref{tab6}, 
the overall performance of the phrase-level code switching is almost the same as that of the word-level code switching, but its performance in cross-lingual transfer is more stable ($\Delta$ decreased by 2.4\%). The bottom row of the table shows the performance of the model without contrastive learning module, F1 decreased by 2.7\% and delta increased by 2.3\% in GIMC(base), suggesting that contrastive learning module of our model does play an important role in the cross-lingual transfer of causal knowledge. 

\subsection{Cross-lingual Transfer Effect of Different Languages}

We plot the performance of zero-shot cross-lingual transfer using different languages as source language in Figure \ref{fig5}. GIMC achieves best cross-lingual performance on Turkish, due to more data and possibly greater linguistic transferability. The large gap between the performance of model on Spanish in monolingual setting and cross-lingual setting leads to its high fluctuation. The drop on Danish much smaller than on other languages, e.g. da-en, which is partly attribute to the effectiveness of our cross-language module, and partly may because of the features of the language such as language families (both Danish and English belong to the Germanic language family). Due to its distinct morphology and syntax compared to the other four languages, Urdu exhibits below-average performance, warranting further attention in future research.


\section{Related Work}
\noindent\textbf{Event Causality Identification} \quad 
Most event causality identification (ECI) methods focus on sentence-level, and the corresponding datasets are mainly English corpora. \citet{ijcai2020p499} proposes a mention masking generalization mechanism to learn event-agnostic, context-specific patterns. \citet{zuo-etal-2021-improving} designs a self-supervised framework to learn context-specific causal patterns from causal statements. However, these sentence-level models fail to predict the causality expressed by multiple sentences. So \citet{gao-etal-2019-modeling} leverages Integer Linear Programming (ILP) to model document-level structures for ECI. \citet{tran-phu-nguyen-2021-graph} constructs several document-level interaction graphs and uses GCN to capture relevant context information. \citet{chen-etal-2022-ergo} proposes a concise graph network with event pairs as nodes to model the transitivity of causation.

\noindent\textbf{Cross-lingual Transfer Learning} \quad Recently, cross-lingual contextualized embeddings have achieved promising results.
e.g., mBERT \cite{devlin-etal-2019-bert}, XLMR \cite{conneau-etal-2020-unsupervised}. Further studies also consider aligning representations between source and target languages during fine-tuning. \citet{liu2020attention} proposes task-related parallel word pairs to generate code-switching sentences for learning the interlingual semantics across languages. \citet{ijcai2020p533} leverage a data augmentation framework to generate multi-lingual codeswitching data to fine-tune mBERT. \citet{qin2022gl} proposes a global-local contrastive learning framework for cross-lingual spoken language understanding.

\section{Conclusion}

In this paper, we propose Heterogeneous Graph Interaction Model with Multi-granularity Contrastive Transfer Learning (GIMC) to simultaneously address document-level and zero-shot cross-lingual ECI. We design a multi-granularity contrastive transfer learning module to align causal representations across language. We construct a heterogeneous graph interaction network to model long-distance dependencies between events. Experiments on Multilingual dataset show the effectiveness of our method in both monolingual and multilingual scenarios. Despite extensive training on high-resource language English, GPT-3.5 with few-shot prompts fails to achieve plug-and-play functionality in cross-lingual task, it appears more practical to develop smaller task-specific models for complex multilingual problems.

\section{Limitations}
This paper are towards the task of zero-shot cross-lingual document-level event causality identification. This work has certain limitations. Firstly, due to lack of dataset, we only conduct experiments on MECI, which is the largest multilingual event causality identification dataset by far, however, the dataset only contain five languages across diverse typologies. It would be beneficial to test the effectiveness of our method on other low-resource language in the future. Secondly, limited by computation resources, our experiments only employ the language-specific PLMs as previous work and the base versions of multilingual PLMs, mBERT and XLMR. Different pre-trained language model as encoder could produce different results. Thirdly, constrained by the high computational cost of API calls, we only conducted analysis on GPT-3.5 with few-shot prompts, thus unable to provide comparisons with other popular LLMs such as GPT-4 \cite{openai2023gpt4}. Furthermore, while we crafted task-specific prompts for GPT-3.5, it could be intriguing to explore additional prompts for testing purposes. We expect future research to encompass a broader range of languages and models.

\section{Acknowledgements}
This work is supported by the National Key Research and Development Program of China (No. 2022ZD0160503), the National Natural Science Foundation of China (No. 62176257 ). This work is also supported by the Youth Innovation Promotion Association CAS.

\nocite{*}
\section{Bibliographical References}\label{sec:reference}

\bibliographystyle{lrec-coling2024-natbib}
\bibliography{lrec-coling2024-example}

\section{Language Resource References}
\label{lr:ref}
\bibliographystylelanguageresource{lrec-coling2024-natbib}
\bibliographylanguageresource{languageresource}

\appendix
\section{Dependency Relations}
\label{appendix A}
The table \ref{tab7} lists the 19 dependency relations we used in this paper. The upper part of the table follows the main organizing principles of the UD taxonomy such that rows correspond to functional categories in relation to the head while columns correspond to structural categories of the dependent. The lower part of the table lists relations that are not dependency relations in the narrow sense.

\begin{table}[!ht]
\centering
\renewcommand\arraystretch{1.25}
\resizebox{\linewidth}{!}{
\begin{tabular}{c|c|c|c}
\toprule

\multicolumn{1}{l}{}\vline & \textbf{Nominals} & \textbf{Clauses} & \textbf{Modifier words} \\

\midrule

\multicolumn{1}{l}{\multirow{4}{*}{\textbf{Core arguments}}} \vline & nsubj & \multicolumn{1}{c}{\multirow{4}{*}{csubj}} \vline& \multicolumn{1}{l}{\multirow{4}{*}{}} \\
\multicolumn{1}{l}{}\vline & nsubj:pass & \multicolumn{1}{c}{}\vline & \multicolumn{1}{l}{}  \\
\multicolumn{1}{l}{}\vline & obj & \multicolumn{1}{c}{} \vline& \multicolumn{1}{l}{}  \\
\multicolumn{1}{l}{}\vline & iobj & \multicolumn{1}{c}{} \vline& \multicolumn{1}{l}{}  \\

\midrule

\multicolumn{1}{l}{\multirow{5}{*}{\textbf{Non-core dependents}}}\vline & obl & \multicolumn{1}{c}{\multirow{5}{*}{advcl}}\vline & \multicolumn{1}{c}{\multirow{5}{*}{advmod}}   \\
\multicolumn{1}{l}{}\vline & obl:loc & \multicolumn{1}{c}{}\vline & \multicolumn{1}{c}{}  \\
\multicolumn{1}{l}{}\vline & obl:tmod & \multicolumn{1}{c}{} \vline& \multicolumn{1}{c}{}  \\
\multicolumn{1}{l}{}\vline & obl:npmod & \multicolumn{1}{c}{} \vline& \multicolumn{1}{c}{} \\
\multicolumn{1}{l}{}\vline & dislocated & \multicolumn{1}{c}{} \vline& \multicolumn{1}{c}{} \\

\midrule

\multicolumn{1}{l}{\multirow{2}{*}{\textbf{Nominal dependents}}}\vline & \multicolumn{1}{c}{\multirow{2}{*}{appos}} \vline& acl & \multicolumn{1}{c}{}  \\

\multicolumn{1}{c}{}\vline & \multicolumn{1}{c}{}\vline & acl:relcl & \multicolumn{1}{c}{}  \\

\bottomrule
\toprule

\multicolumn{1}{c}{\textbf{Coordination}}\vline & \multicolumn{2}{c}{\multirow{1}{*}{\textbf{Loose}}}\vline & \textbf{Other} \\

\midrule

\multicolumn{1}{c}{\multirow{2}{*}{conj}}\vline & \multicolumn{2}{c}{\multirow{1}{*}{list}}\vline & \multicolumn{1}{c}{\multirow{2}{*}{root}} \\

\multicolumn{1}{l}{}\vline & \multicolumn{2}{c}{\multirow{1}{*}{parataxis}}\vline & \multicolumn{1}{l}{} \\

\bottomrule
\end{tabular}
}

\caption{\label{tab7}Dependencies with rich semantics that we reserved in this paper}
\end{table}

\begin{figure}
   \centering{ 
   \includegraphics[width=0.49\textwidth]{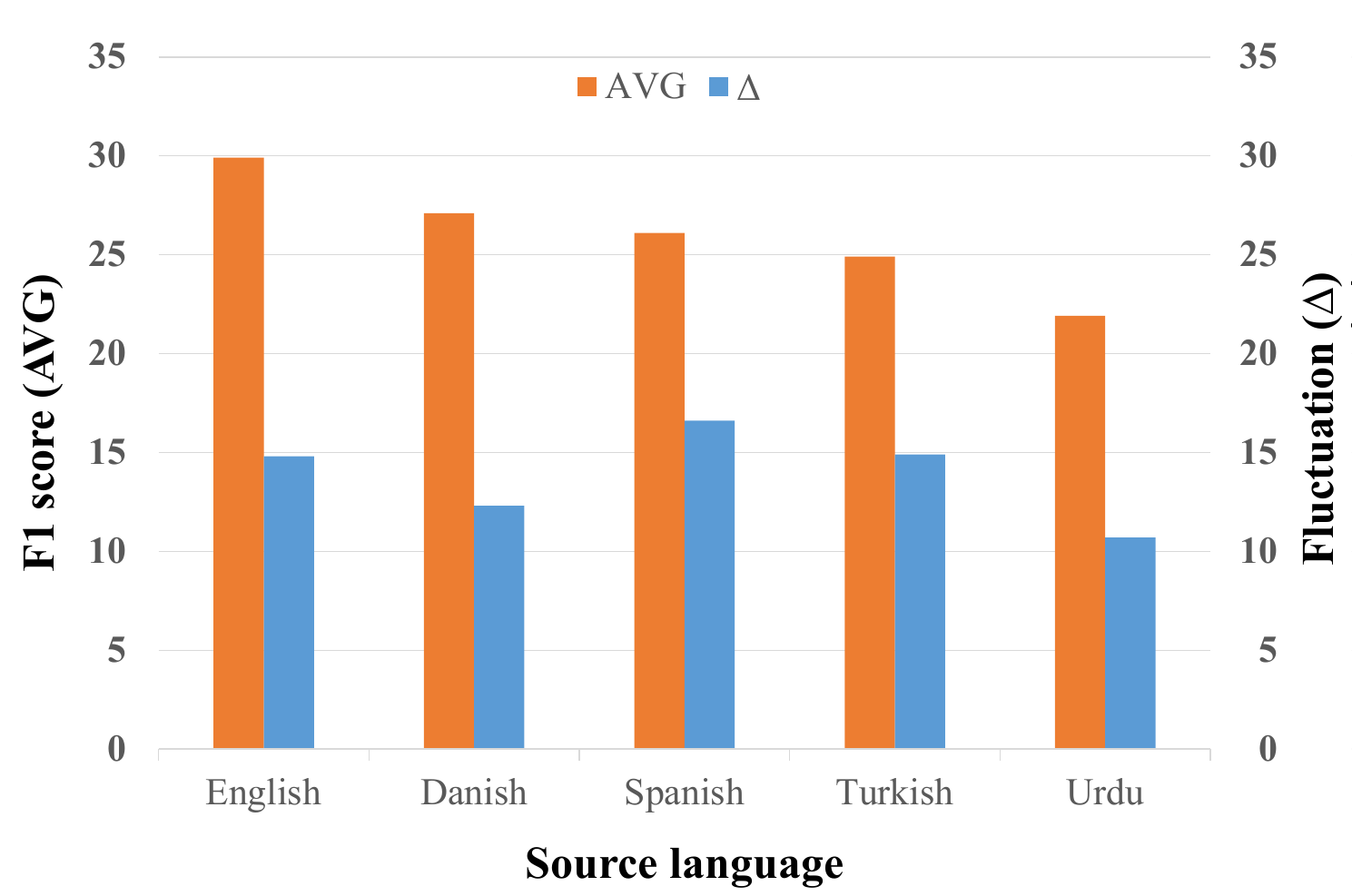}}  
   \caption{Average F1 score(AVG) and fluctuation($\Delta$) of GPT-3.5 with different language specific prompt.}
   \label{fig6}
\end{figure}

\section{Experiment Details of GPT-3.5}
\label{appendix B}

\subsection{Prompts}
\begin{figure*}[htbp]
	 	\centering  
	 	\includegraphics[width=1\textwidth]{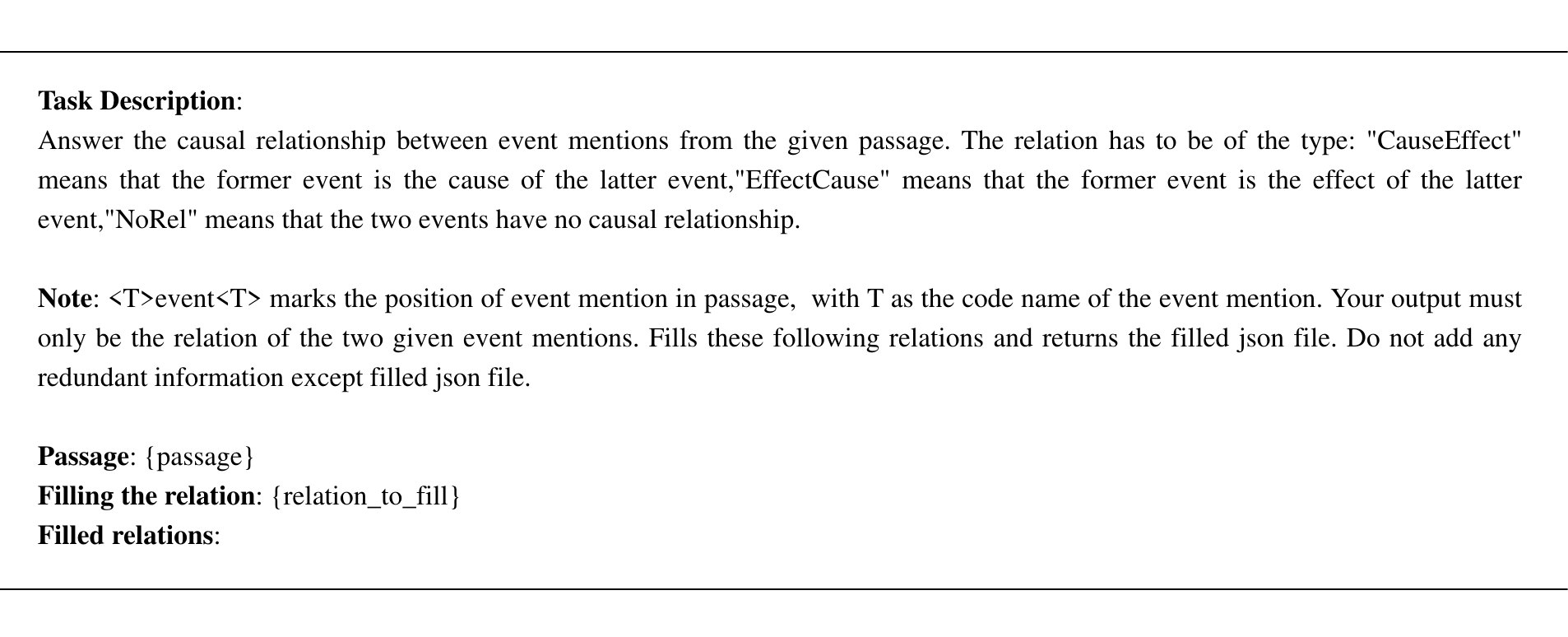}    
	 	\caption{Zero-shot prompt for GPT-3.5}
	 	\label{fig7}
\end{figure*}

In the monolingual setting, we employ a zero-shot prompt as illustrated in Figure \ref{fig7}. We provide a task description and specify the required output format. In the multilingual setting, we have designed five different language-specific few-shot prompts. In each prompt, we provide a document and 13 event pairs, where the events mentioned in the document are marked with special symbols (i.e., <T>). Among the 13 event pairs, 8 are causal event pairs, and 5 are unrelated event pairs.

\subsection{Analysis}

Models were queried through the OpenAI API between December 2022 and May 2023. We illustrate the multilingual performance of GPT-3.5 in Figure \ref{fig6}. It can be observed that GPT-3.5 performs the best in English and the worst in Urdu. This indicates that GPT-3.5 has insufficient understanding of languages other than English, especially low-resource language. Moreover, even in the case of high-resource language English, the overall performance of GPT-3.5 is below 30\%. In addition to the possibility of imperfect prompt design, this outcome is highly likely to be attributed to the complexity of the task and the differences between languages.

\begin{figure*}[htbp]
	 	\centering  
	 	\includegraphics[width=1\textwidth]{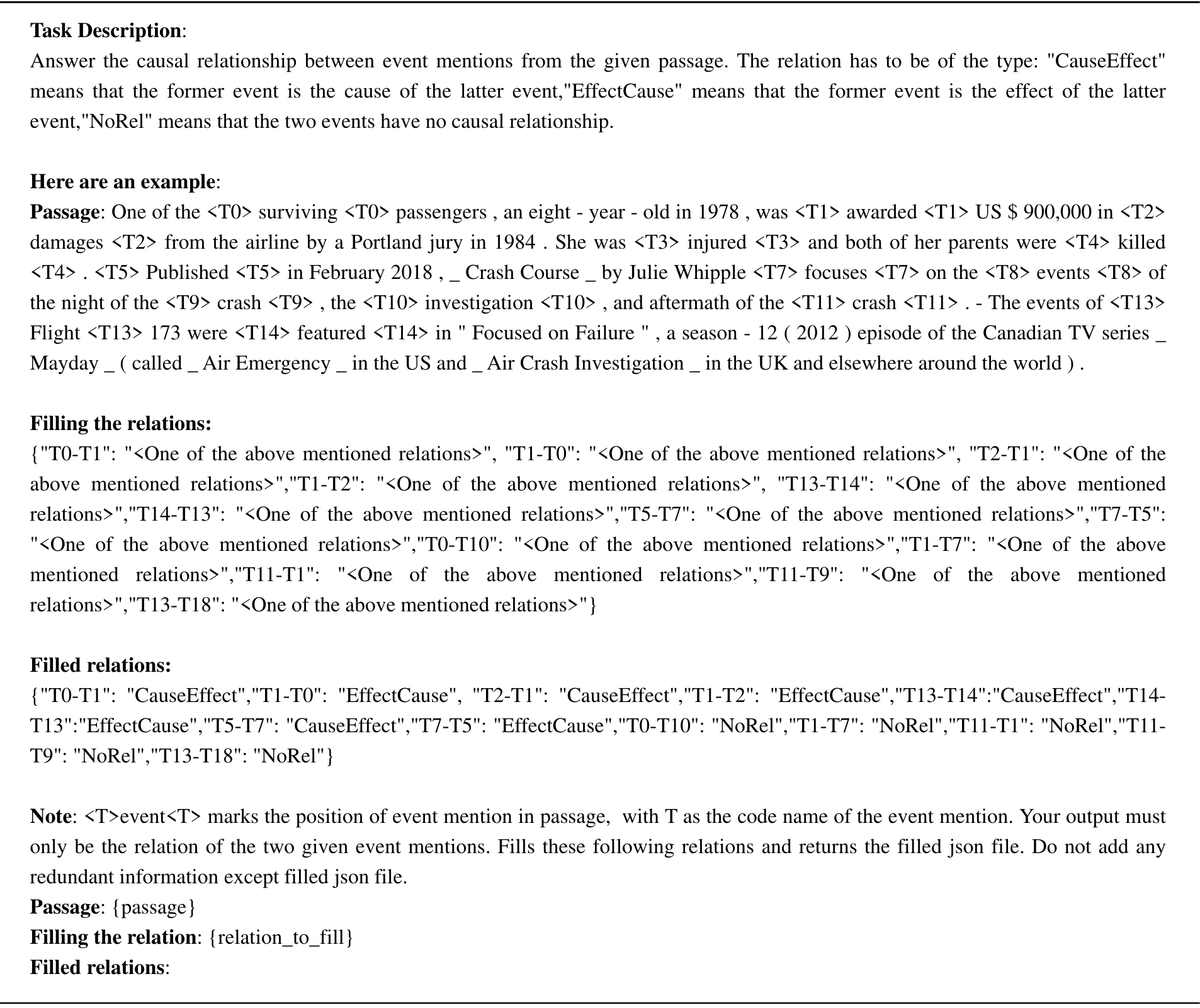}    
	 	\caption{English few-shot prompt}
	 	\label{fig8}
\end{figure*}

\begin{figure*}[htbp]
	 	\centering  
	 	\includegraphics[width=1\textwidth]{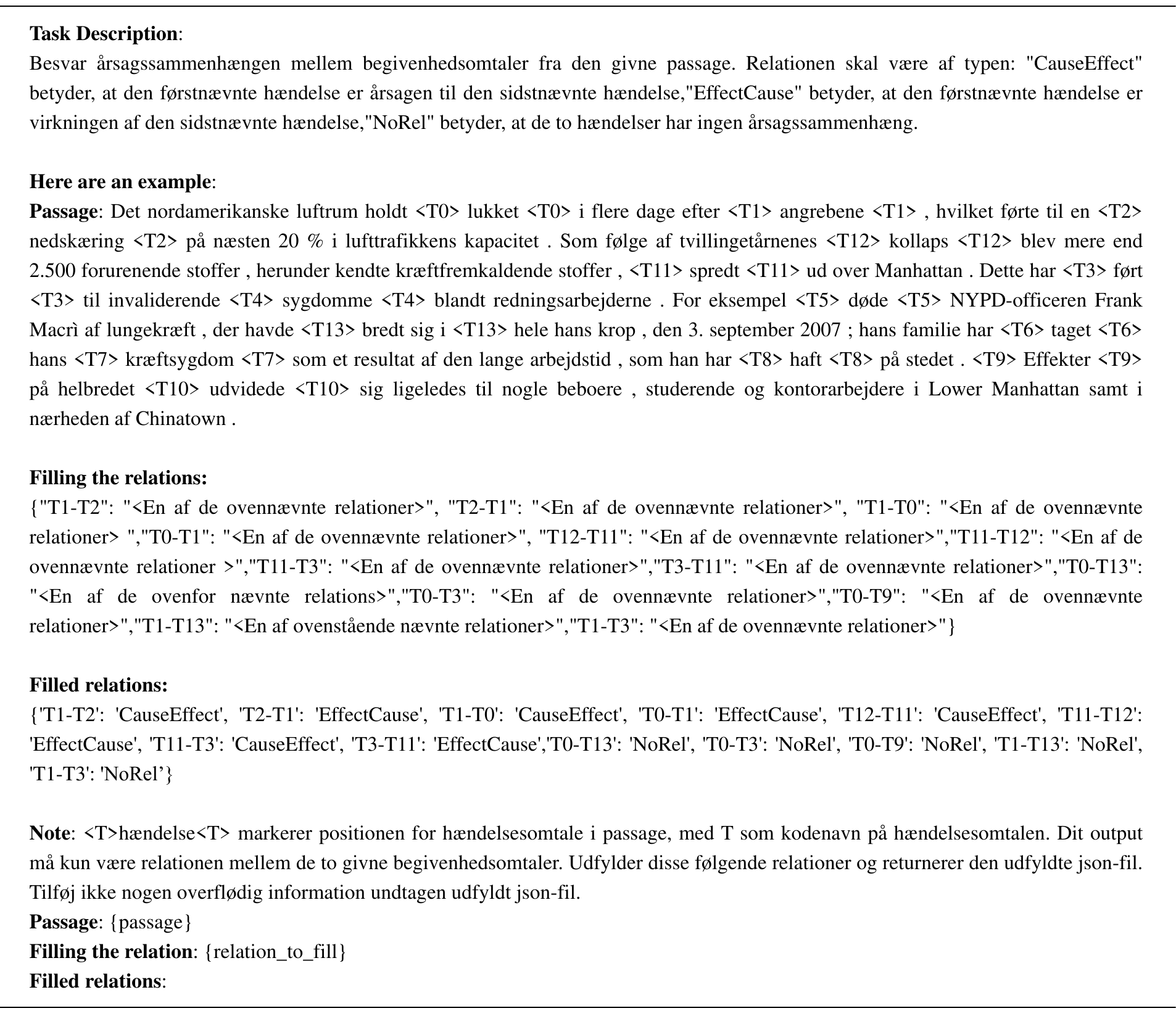}    
	 	\caption{Danish few-shot prompt}
	 	\label{fig9}
\end{figure*}

\begin{figure*}[htbp]
	 	\centering  
	 	\includegraphics[width=1\textwidth]{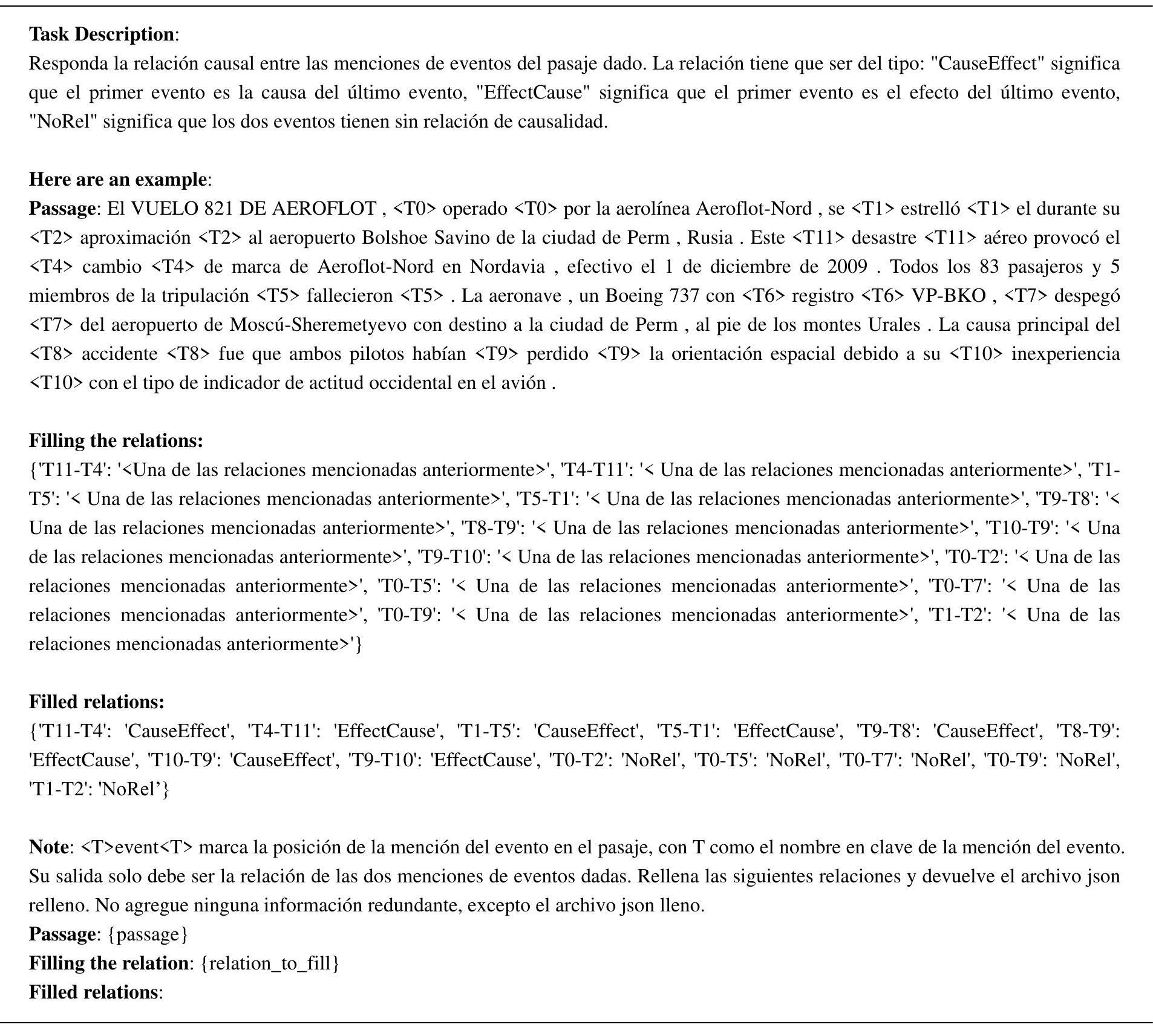}    
	 	\caption{Spanish few-shot prompt}
	 	\label{fig10}
\end{figure*}

\begin{figure*}[htbp]
	 	\centering  
	 	\includegraphics[width=1\textwidth]{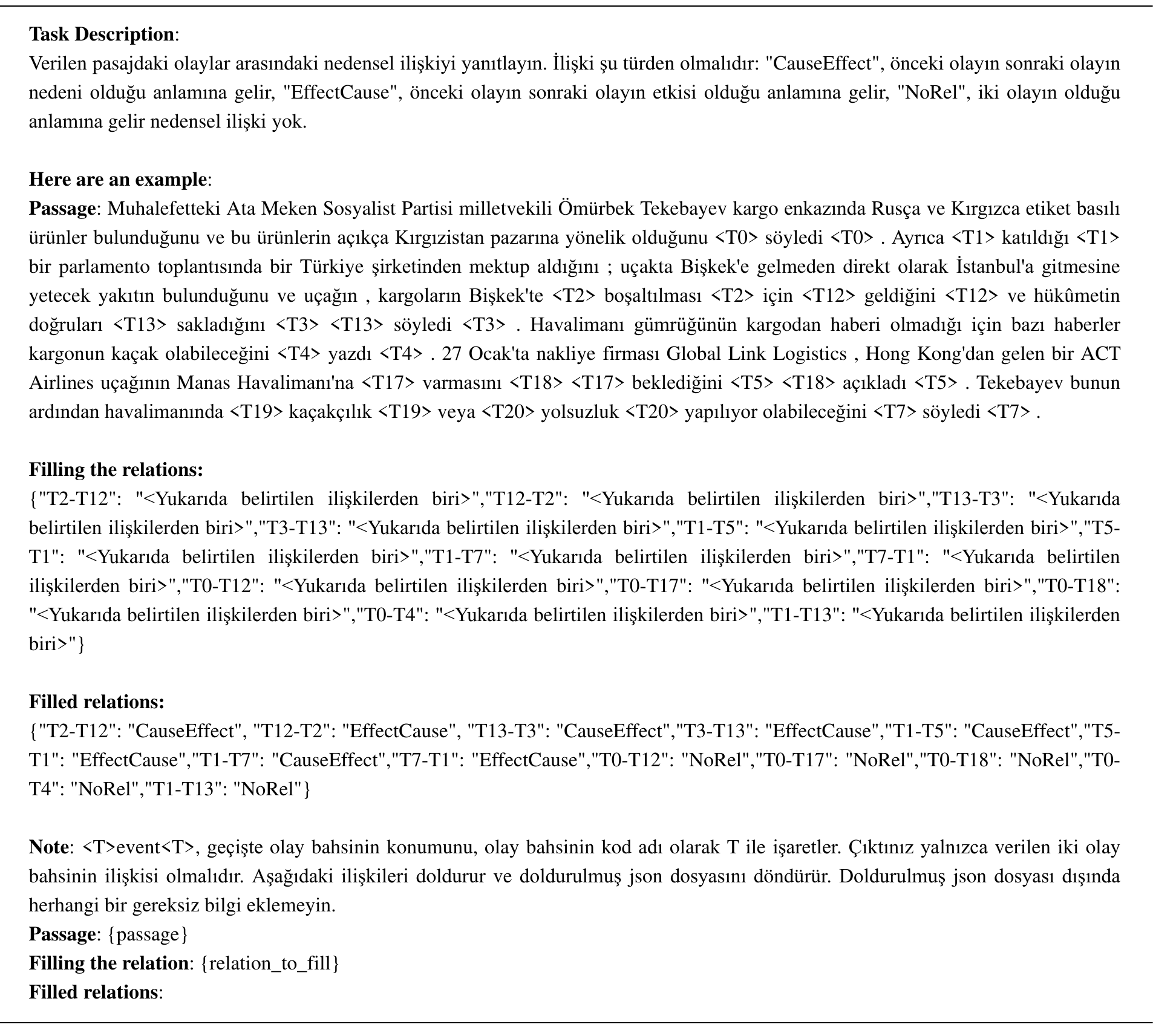}    
	 	\caption{Turkish few-shot prompt}
	 	\label{fig11}
\end{figure*}

\begin{figure*}[htbp]
	 	\centering  
	 	\includegraphics[width=1\textwidth]{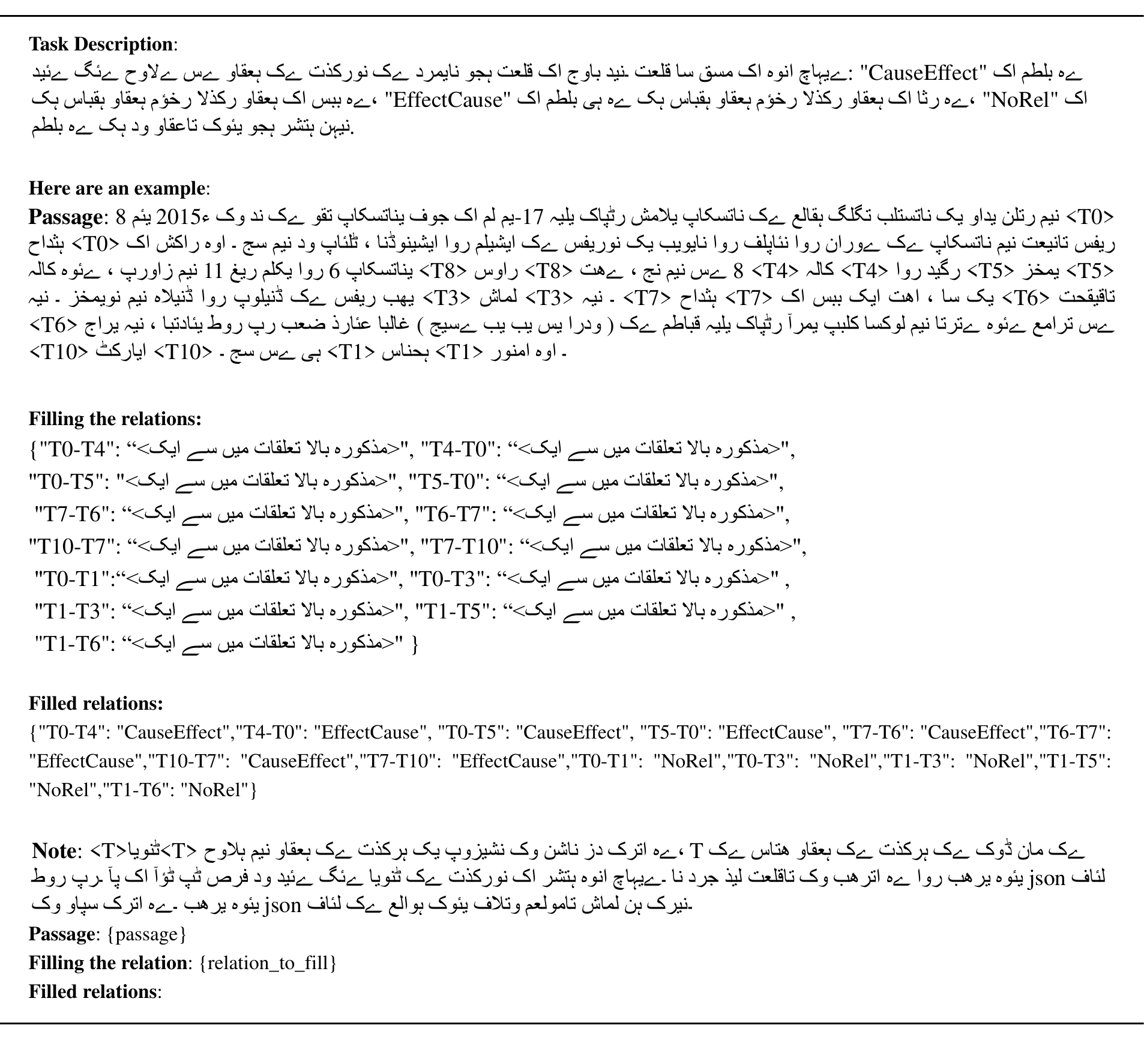}    
	 	\caption{Urdu few-shot prompt}
	 	\label{fig12}
\end{figure*}

\end{document}